\documentclass[lettersize,journal]{IEEEtran}
\usepackage{amsmath,amsfonts}
\usepackage{graphicx}
\usepackage{cite}
\usepackage{picinpar}
\usepackage{amsmath}
\usepackage{mathtools}
\usepackage{url}
\usepackage{flushend}
\usepackage[latin1]{inputenc}
\usepackage{colortbl}
\usepackage{soul}
\usepackage{multirow}
\usepackage{pifont}
\usepackage{color}
\usepackage{alltt}
\usepackage[hidelinks]{hyperref}
\usepackage{enumerate}
\usepackage{siunitx}
\usepackage{epstopdf}
\usepackage{pbox}
\usepackage{overpic}
\usepackage[table]{xcolor}
\usepackage{algorithmic}
\usepackage{algorithm}
\usepackage{makecell}
\usepackage{soul}
\usepackage{threeparttable}
\usepackage{amssymb}
\makeatletter

\newcommand{\Rmnum}[1]{\expandafter\@slowromancap\romannumeral #1@}
\makeatother

\def\etal{{\em et al.~}}
\def\ie{\emph{i.e.,~}}

\newcommand{\sorb}[1]{{\textcolor[rgb]{0.72,0.00,0.00}{\textbf{#1}}}}

\newcommand{\figref}[1]{Fig.~\ref{#1}}
\newcommand{\tabref}[1]{Table.~\ref{#1}}

\newcommand{\secref}[1]{Sec.~\ref{#1}}

\soulregister{\figref}7
\soulregister{\cite}7
\soulregister{\tabref}7
\soulregister{\etal}7
\soulregister{\Rmnum}7
\soulregister{\sorb}7

\newlength\savedwidth
\newcommand{\whline}[1]{\noalign{\global\savedwidth\arrayrulewidth \global\arrayrulewidth #1}%
                   \hline \noalign{\global\arrayrulewidth\savedwidth}}

\begin{document}
\title{\textbf{Phase-Based Multi-Gait Learning for a Salamander-Like Robot}}

\author{
\vspace{0.3cm}
\fontsize{13pt}{16pt}\selectfont Zhiang Liu$^*$\quad Yang Liu$^*$\quad Yongchun Fang$^{\dag}$\quad Xian Guo \\ \textbf{Nankai University}
\thanks{$^*$ Equal contribution; order determined by coin flip.}
\thanks{$^\dag$ Corresponding author.}}

\markboth{}%
{Shell \MakeLowercase{\textit{et al.}}: A Sample Article Using IEEEtran.cls for IEEE Journals}

\IEEEpubidadjcol

\maketitle
	
\begin{abstract}
Salamander-like robots are designed inspired by the skeletal structure of their biological counterparts. However, existing controllers cannot fully exploit these morphological features and largely rely on predefined patterns or joint trajectories, which prevents the generation of diverse and flexible gaits and limits their applicability in real-world scenarios.
In this paper, we propose a phase-based learning framework that enables the robot to acquire a diverse repertoire of gaits without using reference motions. Each body part is controlled by a phase variable capable of forward and backward evolution, with a phase coverage reward to promote the exploration of the leg phase space.
Additionally, morphological symmetry of the robot is incorporated via data augmentation, improving sample efficiency and enforcing both motion-level and task-level symmetry in learned behaviors. 
Extensive experiments show that the robot successfully acquires 22 representative gaits exhibiting both dynamic and symmetric movements, demonstrating the effectiveness of the proposed learning framework.
\end{abstract}


\definecolor{limegreen}{rgb}{0.2, 0.8, 0.2}
\definecolor{forestgreen}{rgb}{0.13, 0.55, 0.13}
\definecolor{greenhtml}{rgb}{0.0, 0.5, 0.0}

\section{Introduction}
Salamander-like robots are designed to replicate the body structure of their biological counterparts, exhibiting characteristics such as a low center of gravity and high stability, which make them promising candidates for real-world applications like disaster relief~\cite{karakasiliotis2016cineradiography} and environmental monitoring~\cite{horvat2015inverse}. However, while existing methods~\cite{liu2025diffusion, liu2026autonomous} have improved the adaptability of these robots in complex environments, their resulting gait patterns are often rigid and limited. Prior research on legged gait control~\cite{yang2022fast,fu2022minimizing,zhu2022generic,saputra2022combining} has shown that the optimal gait may vary with speed, terrain, or task context, underscoring the importance of enabling robots to autonomously discover and master a richer repertoire of locomotor strategies to further enhance performance in higher-level tasks.

Early efforts toward gait control in salamander-like robots predominantly rely on manually designed methods. A common approach~\cite{horvat2017spine} is to predefine specific gait patterns, while another draws inspiration from biological central pattern generators (CPGs)~\cite{crespi2013salamandra,haomachai2024transition}, which generate rhythmic joint trajectories online and enable smooth transitions between different movement modes through parameter modulation. However, such CPG-based or other manually designed frameworks inherently depend on expert knowledge, and their enforced structural constraints still limit the diversity of the motion behaviors that can be expressed.

It is instructive to draw inspiration from recent advances in other legged robots, which no longer rely on predefined patterns or trajectories as direct outputs. Instead, reference information is integrated into the controller input---along with user commands and environmental feedback---to guide the generation of more flexible behaviors. For instance, some methods train end-to-end (E2E) neural networks to directly output joint angles and shape the learning objective by comparing the resulting motion patterns with reference foot contacts~\cite{bellegarda2025allgaits,tang2023saytap} or trajectories~\cite{shao2021learning,liu2024skill}, enabling the robot to acquire multiple gait styles through reinforcement signals. 
To achieve more life-like agility, many studies utilize motion capture data or recorded videos of animals and humans as reference motions, either through MPC-based planners that optimize reference trajectories subsequently tracked by a whole-body controller~\cite{kang2021animal,kang2022animal}, or through imitation learning techniques~\cite{zare2024survey} that acquire locomotion skills from demonstrations~\cite{bin2020learning,bohez2022imitate,han2024lifelike}.
While these approaches have improved the diversity of gaits to some extent, the learned skills are often confined to a limited set of reference patterns and lack generalization to unseen conditions. Moreover, for many bio-inspired prototypes like salamanders, suitable motion datasets are simply unavailable, which further constrains the scalability of such frameworks.

Alternatively, some learning-based studies have attempted to acquire gait controllers without any references. For example, unsupervised skill discovery methods aim to learn diverse motion patterns by maximizing mutual information (MI) between skills and states~\cite{sharma2020emergent,sharmadynamics,laskin2022contrastive,liu2025balancing}. However, the behaviors obtained in this manner often exhibit low motion quality---being disordered, uncoordinated, or physically unnatural---which makes them difficult to transfer to real robotic systems.
An inspiring line of work~\cite{bellegarda2022cpg,bellegarda2024visual} integrates reinforcement learning (RL) with a decoupled CPG model by treating the phase as the policy observation and the phase velocity as the action. This formulation produces stable and adaptive gaits, yet the behavioral diversity remains limited, as the phase velocity is restricted to positive values.

Another challenge in controlling salamander-like robot gait is the uneven exploration of the state-action space. When trained under diverse translational and rotational commands, the policy may perform efficiently in some directions but converge to local optima in others, leading to unbalanced behavior. A promising solution is to exploit the robot's morphological symmetries, which recent studies~\cite{apraez2025morphological} have shown to serve as powerful physics-informed priors for improving both analytical and learning-based control.
Building on this idea, subsequent studies~\cite{suleveraging,mittal2024symmetry} have applied symmetry-based formulations to practical quadruped tasks such as obstacle climbing and door opening, showing improved learning efficiency and motion consistency. However, discussions and analyses focused on symmetric gait patterns remain limited.

In this work, we propose a phase-based multi-gait learning framework for a salamander-like robot that acquires a large repertoire of locomotion gaits under diverse translational and rotational movement commands without using reference motion trajectories.
Each body part of the robot is individually controlled through a phase variable that governs its rhythmic motion. A key feature of our approach is that the action space allows both positive and negative values, making it possible for limbs to move in opposite directions and thereby greatly expanding the range of possible action combinations.
Meanwhile, a reward function based on distance traveled and turning angles within each gait cycle guides the agent to adjust phase velocities in real time.  Furthermore, we incorporate a phase coverage term in the reward function to encourage dynamic motion patterns, preventing the emergence of static postures or unnatural artifacts.
The second key component incorporates the robot's morphological symmetry into learning through data augmentation. By applying front-back and left-right reflections together with their composition to both states and actions, the policy is encouraged to maintain approximate equivariance under these transformations.
This design enhances sample efficiency, enforces symmetry in both motion and trajectories, and results in more natural and robust gait patterns across multiple command types.
The main contributions of this work are summarized as follows:
\begin{itemize}
    \item A phase-based multi-gait learning framework is proposed, where each body part is governed by a phase variable, and a phase coverage reward encourages exploration of the leg phase space.
    \item Morphological symmetry is incorporated during learning through data augmentation, promoting motion-level and task-level symmetry in the learned behaviors.
    \item The proposed framework is systematically validated through extensive experiments, enabling the salamander-like robot to acquire 22 representative gaits exhibiting dynamic and symmetric movements.
\end{itemize}

The remainder of this paper is organized as follows.
In \secref{sec:pre}, we introduce the fundamental concepts and notation underlying reinforcement learning, symmetry groups, and symmetric Markov decision processes.
Thereafter, the detailed methodology of the proposed framework is presented in \secref{sec:method}.
\secref{sec:exp} illustrates the results of experiments, where a thorough analysis is provided to elucidate the phenomena.
Finally, the conclusion is drawn in \secref{sec:conclusion}.

\section{Preliminaries}\label{sec:pre}
This section briefly introduces the key concepts and notation, including reinforcement learning, symmetry groups, and symmetric Markov decision processes, which form the foundation for our learning framework.

\subsection{Reinforcement Learning}
The multi-gait learning task of the salamander-like robot is formulated as a Markov decision process (MDP), defined by the tuple $(\mathcal{S}, \mathcal{A}, \mathcal{P}, \mathcal{R})$, where $\mathcal{S}$ denotes the state space, $\mathcal{A}$ is the action space, $\mathcal{P}:\mathcal{S}\times\mathcal{A}\times\mathcal{S}\rightarrow[0,1]$ specifies the transition dynamics $\mathcal{P}(\boldsymbol{s}_{t+1}\mid \boldsymbol{s}_t,\boldsymbol{a}_t)$, and $\mathcal{R}:\mathcal{S}\times\mathcal{A}\times\mathcal{S}\rightarrow\mathbb{R}$ is the reward function. At each timestep, the robot observes its current state $\boldsymbol{s}_t \in \mathcal{S}$, selects an action $\boldsymbol{a}_t \in \mathcal{A}$ according to the policy $\pi_\theta(\boldsymbol{a}_t \mid \boldsymbol{s}_t)$ parameterized by $\theta$, and receives a reward $r_t$ while transitioning to the next state $\boldsymbol{s}_{t+1}$. The objective of RL is to learn an optimal policy that maximizes the expected return:
\begin{equation}
\label{rewardE}
\pi^* = \underset{\pi}{\arg\max} \; \mathbb{E}\left[\sum_{t=0}^{\infty} \gamma^t r_t\right],
\end{equation}
where 
$\mathbb{E}[\cdot]$ denotes the expectation operator and 
$\gamma \in [0,1)$ is the discount factor.

\subsection{Symmetry Groups}
Many robots with repetitive or bilateral structures, such as salamander-like robots, exhibit inherent morphological symmetries. We formalize these properties using group theory, which treats symmetry transformations as abstract mathematical objects independent of specific systems~\cite{suleveraging,apraez2025morphological}.

A symmetry transformation is an invertible mapping that preserves a fundamental property of an object. The set of all such transformations constitutes a symmetry group, denoted as 
$\mathbb{G}:= \{ e, g_1, g_1^{-1}, g_2, \ldots \}$.
This set is closed under the composition operation 
$\circ : \mathbb{G} \times \mathbb{G} \rightarrow \mathbb{G}$ 
and the inversion operation 
$(\cdot)^{-1} : \mathbb{G} \rightarrow \mathbb{G}$. 
Specifically, for any $g_1, g_2 \in \mathbb{G}$, 
the composition $g_1 \circ g_2 \in \mathbb{G}$, 
and each transformation $g \in \mathbb{G}$ has an inverse 
$g^{-1} \in \mathbb{G}$ satisfying $g^{-1} \circ g = e$, 
where $e$ denotes the identity transformation. 
The order $|\mathbb{G}|$ of the group represents the number 
of distinct transformations it contains.
Among the various symmetry groups, a particularly fundamental example relevant to our robot is the \emph{reflection group}, denoted as
$\mathbb{G}:=\mathbb{C}_{2}=\{e,g_{s} \,\vert\, g_{s}^2=e\}$, which contains the identity and a reflection transformation $g_{s}$ that is its own inverse, \ie $g_{s}^2:=g_{s}\circ g_{s}=e$.

To formalize the action of the reflection group on the robot's vector spaces, we employ a \emph{group representation}. For a vector space $\mathcal{X} \subseteq \mathbb{R}^n$, a representation of $\mathbb{G}$ is a homomorphism $\rho_\mathcal{X} : \mathbb{G} \rightarrow \mathbb{GL}(\mathcal{X})$, which maps group elements into invertible linear operators on $\mathcal{X}$. This mapping preserves the group structure, implying that for any $g_a, g_b \in \mathbb{G}$, we have $\rho_\mathcal{X}(g_a \circ g_b) = \rho_\mathcal{X}(g_a)\rho_\mathcal{X}(g_b)$, and for any $g \in \mathbb{G}$, $\rho_\mathcal{X}(g^{-1}) = [\rho_\mathcal{X}(g)]^{-1}$. The action of a group element $g \in \mathbb{G}$ on a point $\boldsymbol{x} \in \mathcal{X}$ is then defined as $g \, \triangleright\, \boldsymbol{x} := \rho_\mathcal{X}(g)\boldsymbol{x} \in \mathcal{X}$. A vector space equipped with such an action is referred to as a \emph{symmetric vector space}.

\subsection{Symmetric MDPs}

A function $f: \mathcal{X} \rightarrow \mathcal{Y}$ between two symmetric vector spaces is said to be $\mathbb{G}$-invariant if its output remains unchanged under any group transformation applied to the input, \ie $f(\boldsymbol{x}) = f(\rho_\mathcal{X}(g)\boldsymbol{x})$ for all $g \in \mathbb{G}$, $\boldsymbol{x} \in X$, and $\mathbb{G}$-equivariant if applying a transformation to the input before computing the function is equivalent to applying the transformation to the output after computing the function, \ie $\rho_\mathcal{Y}(g) f(\boldsymbol{x}) = f(\rho_\mathcal{X}(g)\boldsymbol{x})$ for all $g \in \mathbb{G}$, $\boldsymbol{x} \in \mathcal{X}$.

Building on this, an MDP $(\mathcal{S}, \mathcal{A}, \mathcal{P}, \mathcal{R})$ is considered \emph{symmetric} under the group $\mathbb{G}$ if both the state space $\mathcal{S}$ and action space $\mathcal{A}$ are symmetric vector spaces with representations $\rho_\mathcal{S}$ and $\rho_\mathcal{A}$~\cite{suleveraging}, 
and if for all $g \in \mathbb{G}$, $\boldsymbol{s}_{t+1},\boldsymbol{s}_{t} \in \mathcal{S}$, and $\boldsymbol{a}_{t} \in \mathcal{A}$, 
the transition density $\mathcal{P}(g \,\triangleright\, \boldsymbol{s}_{t+1} \mid g \,\triangleright\, \boldsymbol{s}_{t}, g \,\triangleright\, \boldsymbol{a}_{t}) = \mathcal{P}(\boldsymbol{s}_{t+1} \mid \boldsymbol{s}_{t}, \boldsymbol{a}_{t})$, and the reward function $\mathcal{R}(g \,\triangleright\, \boldsymbol{s}_{t}, g \,\triangleright\, \boldsymbol{a}_{t}) = \mathcal{R}(\boldsymbol{s}_{t},\boldsymbol{a}_{t})$ are $\mathbb{G}$-invariant. 
These conditions ensure that the state distribution at any timestep remains $\mathbb{G}$-invariant and imply that the optimal policy $\pi^*$ and the optimal value function $V^{\pi^*}$ satisfy the symmetry constraints $g \,\triangleright\, \pi^*(\boldsymbol{s}_{t}) = \pi^*(g \,\triangleright\, \boldsymbol{s}_{t})$ and $V^{\pi^*}(g \,\triangleright\, \boldsymbol{s}_{t}) = V^{\pi^*}(\boldsymbol{s}_{t})$, which provide strong inductive biases that can be leveraged in RL.

\begin{figure}[tp]
	\centering
\includegraphics[width=1\linewidth]{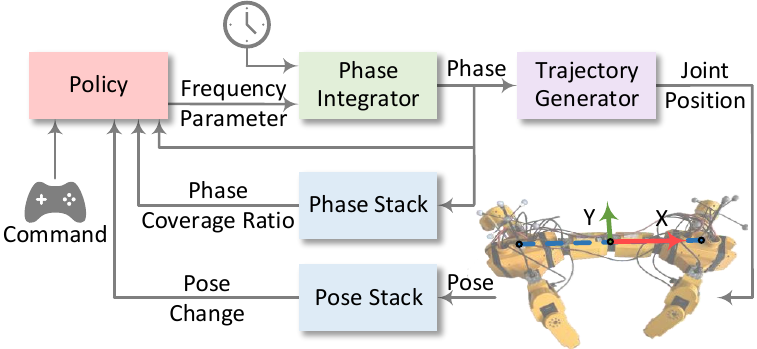}
	\caption{Overview of the learning framework, including the robot prototype and the body-fixed coordinate system. The robot has 15 actuated degrees of freedom in total, consisting of a segmented spine with 3 DOFs that allows horizontal bending and four legs, each with 3 joints. The origin of the body coordinate system is located at the midpoint between the front and hind girdles, and the x-axis points from the hind girdle toward the front girdle.}
	\label{fig:framework}
\end{figure}

\section{Methodology}\label{sec:method}

In this section, we first provide an overview of the proposed phase-based multi-gait learning framework. The overall architecture is illustrated in \figref{fig:framework}. The policy outputs frequency parameters that are integrated over time to obtain the phases, which in turn drive the trajectory generator to produce joint positions for the salamander-like robot. The phases are also recorded in the phase stack for computing phase coverage ratios and fed back to the policy as additional inputs, while the robot's poses are stored in the pose stack to estimate pose changes. The policy thus receives command signals, phases, phase coverage ratios, and pose changes as inputs to guide action selection. The key components of this framework are introduced below.

\subsection{Action Space}
\begin{figure}[tp]
	\centering
\includegraphics[width=1\linewidth]{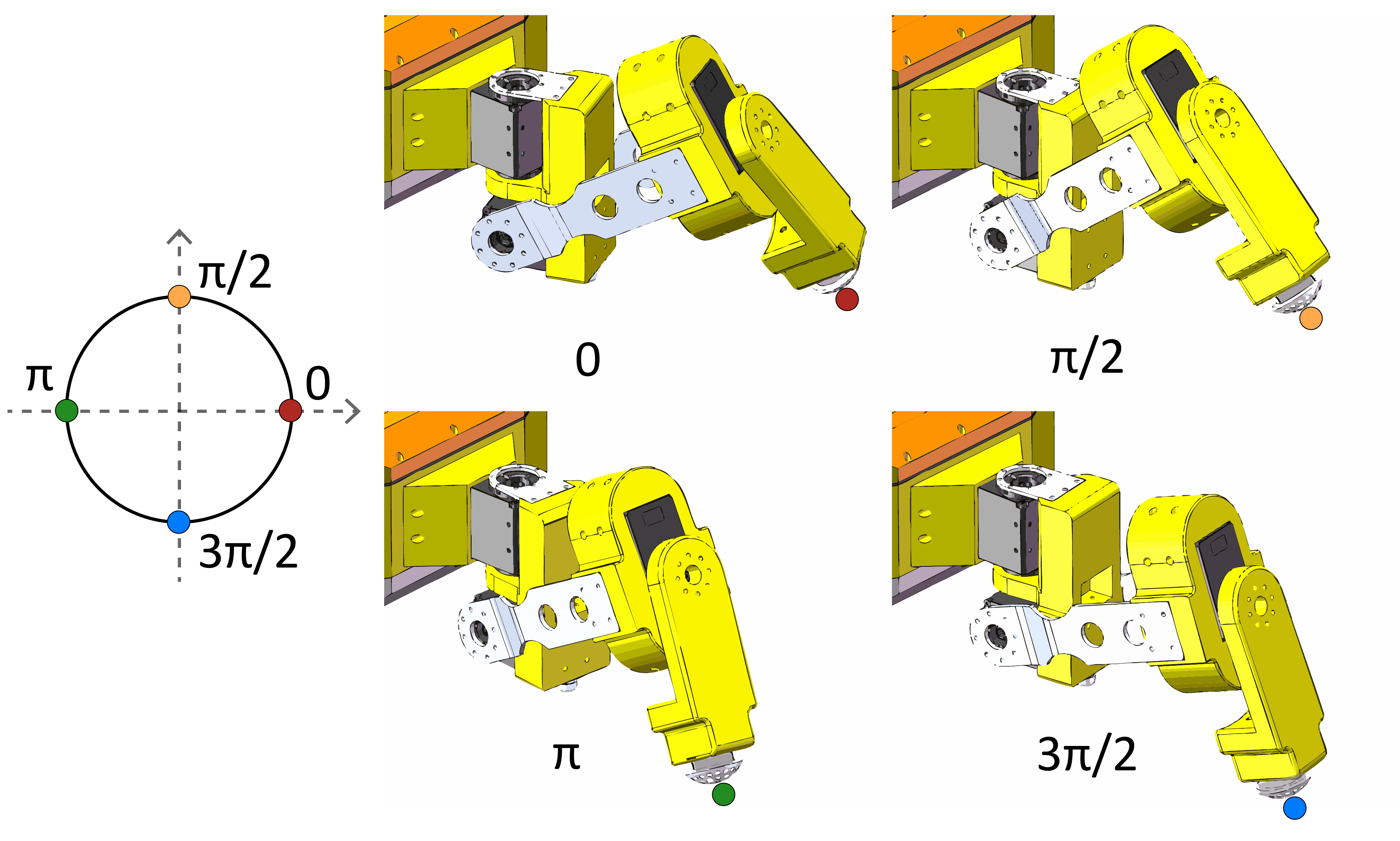}
	\caption{Mapping from phase variables to leg joint angles. Colors represent different phase values, where the leg is in the swing phase for \([0,\pi)\) and in the stance phase for \([\pi,2\pi)\).
     }
	\label{fig:tg}
\end{figure}

Each of the robot's five body parts (left-front (LF), left-hind (LH), right-front (RF), right-hind (RH) legs, and the spine (S)) is controlled independently by a phase variable. The trajectory generator converts the phase variables ${\phi_{\mathrm{LF}}, \phi_{\mathrm{LH}}, \phi_{\mathrm{RF}}, \phi_{\mathrm{RH}}, \phi_{\mathrm{S}}}$ defined on the interval \([0,2\pi)\) into joint angles and produces the desired motor positions $\boldsymbol{u}_{\mathrm{tg}} \in \mathbb{R}^{15}$. For each leg $j \in \{\text{LF},\text{LH},\text{RF},\text{RH}\}$, the joint angles are specified as
\begin{align}
\theta_{j,1} &= A \cos(\phi_j), \\
\theta_{j,2} &= A \sin(\phi_j) + \theta_\mathrm{offset}, \\
\theta_{j,3} &= A \cos(\phi_j),
\end{align}
where $A$ denotes a constant amplitude, and $\theta_\mathrm{offset}$ represents a fixed offset providing a baseline posture for the second joint. 
Given the robot's short leg links and limited workspace, a joint-space sinusoidal generator is adopted, as more complex parameterizations are unlikely to provide significant performance benefits.
A sample visualization of the leg positions at different phases is shown in \figref{fig:tg}.  
The spine joints are defined as:
\begin{equation}
\theta_{\text{S},k} = A \cos(\phi_\text{S}), \quad k \in \{1,2,3\}.
\end{equation}
This formulation produces a 15-dimensional target vector
\begin{equation}
\boldsymbol{u}_{\mathrm{tg}} = 
[\theta_{\text{LF},1}, \theta_{\text{LF},2}, \theta_{\text{LF},3}, \ldots, \theta_{\text{S},1}, \theta_{\text{S},2}, \theta_{\text{S},3}]^\top,
\end{equation}
which is subsequently tracked by Proportional-Integral-Derivative controllers.

Unlike recent approaches that apply RL to directly optimize joint angles or the parameters of predefined trajectories, 
our method focuses exclusively on modulating the phase evolution process. 
The phase variables are updated at each control step according to
\begin{equation}
\phi_i \leftarrow \left( \phi_i + \omega_i \cdot \Delta t \right) \bmod 2\pi, \quad i \in \{\text{LF}, \text{LH}, \text{RF}, \text{RH}, \text{S}\},
\end{equation}
where $\omega_i \in \left[-\frac{\pi}{2},\,\frac{\pi}{2}\right]\ \text{rad/s}$ denotes the phase velocity, 
and $\Delta t$ is the control interval.
Phase velocity is allowed to be positive or negative, indicating forward or reverse phase progression. Accordingly, the action space is summarized as the phase velocity vector output by the policy:
\begin{equation}
\boldsymbol{a}_t = [\omega_\text{LF}, \omega_\text{LH}, \omega_\text{RF}, \omega_\text{RH}, \omega_\text{S}].
\end{equation}
By defining this action space, the robot is not constrained to manually designed gaits. 
Instead, the agent gains the flexibility to develop its own coordination strategy among limbs and spine, providing clear interpretability while 
enabling the emergence of adaptive and potentially novel locomotor behaviors.

\subsection{Observation Space}
The state vector $\boldsymbol{s}_t$ consists of task-level commands, recent body pose changes, phase encodings of the limbs and trunk, and per-leg phase coverage statistics.
Specifically, the command vector is defined as $[ v^{*},\omega^{*}  ] $. The first part is the target horizontal direction in the robot's body frame $v^{*}:= [ \cos\psi^{*}, \sin \psi^{*}] =[v_x^{*}, v_y^{*}]  $, where $\psi^{*}\in[0,2\pi)$ denotes the yaw angle of the commanded direction. $ v_x^{*}$ and $v_y^{*}$ are components along the x- and y- axes. When no translation command is given, this term is defined as $[0,0]$. The second part is the turning direction $\omega^{*}\in\{-1,0,1\}$, corresponding to clockwise rotation, no rotation, and counterclockwise rotation, respectively.
Based on this formulation, we select 22 representative commands that comprehensively cover both translational and rotational movements. These commands, summarized in \tabref{tab:gait_type}, are organized into seven categories to form a diverse repertoire for training and evaluation.

During locomotion, the robot's instantaneous linear and angular velocities fluctuate significantly, making them unsuitable for tracking. 
Instead, we measure the traveled distance and the turned angle over a given time interval $T^*$, assuming that the effective learned gaits are $T^*$-periodic. At each timestep, the robot's pose in the world frame is recorded in a stack of length $T^*/ \Delta t $.
The body displacement and yaw change $[\Delta x, \Delta y, \Delta \psi]$ are then estimated by transforming the most recent poses into the robot frame of the earliest entry.

The phases of the four legs and the spine are provided as $[\sin(\phi_i), \cos(\phi_i)]$ for each $i \in \{\text{LF}, \text{LH}, \text{RF}, \text{RH}, \text{S}\}$, yielding a smooth representation on the unit circle that avoids
$2\pi$ discontinuities and stabilizes learning.
To evaluate the performance of each leg's motion within a gait cycle, phase coverage is computed from a parallel stack of the same length, as shown in \figref{fig:framework}. The stored phase values are sorted, circular differences (including the wrap-around gap) are calculated. 
The largest gap $d_{\max}$ represents the unvisited portion of the phase space. Subtracting this gap from $2\pi$ gives the total covered range, and normalizing it by 
$2\pi$ yields the per-leg coverage values
$\text{cov}_i \in [0,1]$ for $i \in \{\text{LF}, \text{LH}, \text{RF}, \text{RH}\}$, summarized as
$\mathbf{c}=[\text{cov}_\text{LF}, \text{cov}_\text{LH}, \text{cov}_\text{RF},
\text{cov}_\text{RH}]$.
The complete state vector is therefore 
\begin{equation}
    \boldsymbol{s}_t = [v_x^{{*}}, v_y^{{*}}, \omega^{{*}}, \Delta x, \Delta y, \Delta \psi, \{\sin\phi_i, \cos\phi_i\}_{i=1}^5, \{\text{cov}_i\}_{i=1}^4].
\end{equation}

\begin{table}[tp]
	\centering
    \begin{threeparttable}
	\small
	\renewcommand{\arraystretch}{1.15}
  \setlength\tabcolsep{0.88mm}
  \caption{Repertoire of 22 Representative Gait Commands Across 7 Categories. }
	\label{tab:gait_type}
	\begin{tabular}{l|c|l|c|c} \whline{1pt}
        Category & No. & Gait & $\psi^*$ & $\omega^*$ \\ \whline{0.7pt}
        \multirowcell{2}[0pt][l]{Linear \\ Locomotion} 
        & 1 & Forward & 0 & 0 \\
        & 2 & Backward & $\pi$ & 0 \\
        \whline{0.7pt}

        \multirowcell{2}[0pt][l]{Lateral \\ Locomotion} 
        & 3 & Left Lateral & $\pi/2$ & 0 \\
        & 4 & Right Lateral & $3\pi/2$ & 0 \\
        \whline{0.7pt}

        \multirowcell{2}[0pt][l]{Rotational \\ Locomotion} 
        & 5 & Clockwise Rotational & - & -1 \\
        & 6 & Counterclockwise Rotational & - & 1 \\
        \whline{0.7pt}

        \multirowcell{4}[0pt][l]{Curved \\ Locomotion} 
        & 7 & Forward-Clockwise Curved & 0 & -1 \\
        & 8 & Forward-Counterclockwise Curved & 0 & 1 \\
        & 9 & Backward-Clockwise Curved & $\pi$ & -1 \\
        & 10 & Backward-Counterclockwise Curved & $\pi$ & 1 \\
        \whline{0.7pt}

        \multirowcell{4}[0pt][l]{Diagonal \\ Locomotion} 
        & 11 & Forward-Left Diagonal & $\pi/4$ & 0 \\
        & 12 & Backward-Left Diagonal & $3\pi/4$ & 0 \\
        & 13 & Backward-Right Diagonal & $5\pi/4$ & 0 \\
        & 14 & Forward-Right Diagonal & $7\pi/4$ & 0 \\
        \whline{0.7pt}

        \multirowcell{4}[0pt][l]{Oblique \\ Locomotion \\ \Rmnum{1}} 
        & 15 & Forward-Left Oblique & $\pi/8$ & 0 \\
        & 16 & Backward-Left Oblique & $7\pi/8$ & 0 \\
        & 17 & Backward-Right Oblique & $9\pi/8$ & 0 \\
        & 18 & Forward-Right Oblique & $15\pi/8$ & 0 \\
        \whline{0.7pt}

        \multirowcell{4}[0pt][l]{Oblique \\ Locomotion \\ \Rmnum{2}} 
        & 19 & Forward-Left Oblique & $3\pi/8$ & 0 \\
        & 20 & Backward-Left Oblique & $5\pi/8$ & 0 \\
        & 21 & Backward-Right Oblique & $11\pi/8$ & 0 \\
        & 22 & Forward-Right Oblique & $13\pi/8$ & 0 \\       
        \whline{1pt}
	\end{tabular} 
    \begin{tablenotes}
\footnotesize
\item \!\!\!\!\!\!\!\!``-'' indicates no translation command.
\end{tablenotes}
\end{threeparttable}
\end{table}

\subsection{Reward Function}
The reward function aims to guide the agent to follow the task-level commands, as well as to improve the quality of the generated motions. 
Specifically, it rewards motion along the target direction, encourages turning when required, and penalizes movement in undesired directions. Additionally, it promotes exploration of the phase space through coverage of limb oscillations. The reward is then composed of the following terms:
\begin{itemize}
    \item \textbf{Forward motion reward:} Maximizes the traveled distance projected onto the command direction:
    \begin{equation}
        r_\text{toward} = (\Delta x\cdot v_x^{*} +\Delta y\cdot v_y^{*}) \cdot (\alpha + \beta |v_y^{*}|).
    \end{equation}
    
    \item \textbf{Perpendicular motion penalty:} Penalizes movements orthogonal to the desired direction:
    \begin{equation}
        r_\text{perp} = -|\Delta x\cdot (-v_y^{*}) +\Delta y\cdot v_x^{*}| \cdot (\zeta + \epsilon |v_x^{*}|).
    \end{equation}
    
    \item \textbf{Turning reward:} Encourages the agent to turn as fast as possible when $\omega^{{*}}$ is nonzero:
    \begin{equation}
        r_\text{turn} = \delta \, \omega^{*} \, \Delta \psi.
    \end{equation}
    
    \item \textbf{Undesired turn penalty:} Penalizes rotation when $\omega^{{*}}$ is zero:
    \begin{equation}
        r_\text{u\_turn} = -(1 - |\omega^{*}|) \frac{|\Delta \psi|}{2}.
    \end{equation}
    
    \item \textbf{Undesired translation penalty:} Penalizes movement when no translation command is given:
    \begin{equation}
        r_\text{u\_move} = -(1 - \sqrt{(v_x^{*})^2 + (v_y^{*})^2}) \frac{|\Delta x| + |\Delta y|}{2}.
    \end{equation}
    
    \item \textbf{Phase coverage reward:} Rewarding coverage of limb oscillations $\mathbf{c} = \{c_\text{LF}, c_\text{LH}, c_\text{RF}, c_\text{RH}\}$:
    \begin{equation}
        r_\text{coverage} = \text{mean}(\mathbf{c}) + \min(\mathbf{c}),
    \end{equation}
\end{itemize}
where $\alpha, \beta, \zeta, \epsilon, \delta$
are preset parameters.
The total reward is computed as a weighted sum of the above terms, 
with corresponding weights $w_i$ for $i=1,\ldots,6$.

\subsection{Symmetry-Based Data Augmentation}
The policy is trained using Proximal Policy Optimization (PPO)~\cite{schulman2017proximal}.
When applied to gait control tasks, such model-free RL methods often struggle with inefficient exploration and unnatural behaviors in the absence of reference motions, leading to asymmetric and sometimes unstable gaits.
To address these issues, we exploit the inherent morphological symmetry of the robot and introduce symmetry-based data augmentation.

\begin{figure}[tp]
	\centering
\includegraphics[width=1\linewidth]{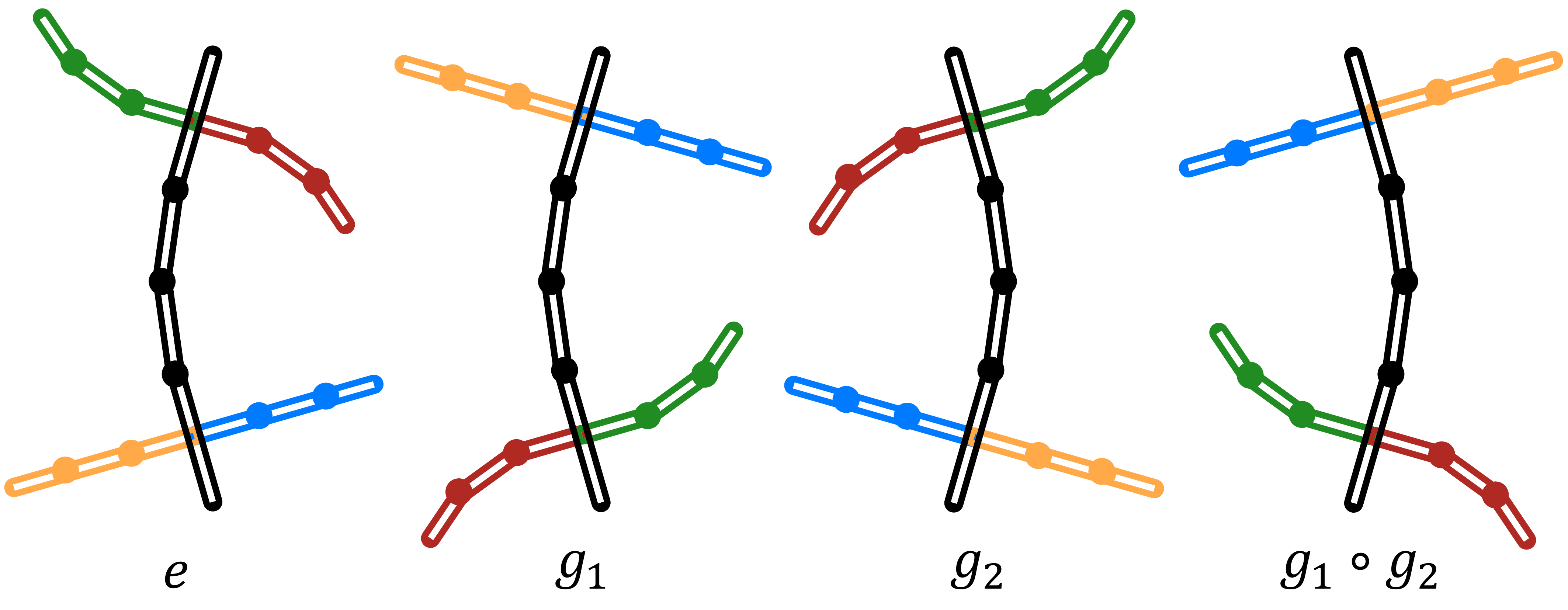}
	\caption{Diagram of the morphological symmetries of the salamander-like robot. Colors indicate phase values consistent with those in \figref{fig:tg}. }
	\label{fig:sym}
\end{figure}

Essentially, morphological symmetries describe how a robot's body structure allows equivalent reconfigurations under certain spatial transformations. As illustrated in \figref{fig:sym}, the salamander-like robot can rearrange the phase relations of its body parts to emulate two orthogonal reflections in 3D space. These transformations can be formally captured by a reflection symmetry group $\mathbb{G}:=\{e,g_{1},g_{2},g_{1}\circ g_{2}\}$, where $e$ denotes the identity transformation, $g_1$ and $g_2$ represent the front-back and left-right reflections, respectively, and $g_1 \circ g_2$ their composition. The corresponding actions of these transformations on the robot's state and action spaces are detailed in \tabref{tab:trans}.
 
During training, we perform augmentation by applying these symmetric transformations to the sampled batches, effectively modeling the control problem as a symmetric MDP that admits the reflection group $\mathbb{G}$. In this formulation, the transition dynamics and reward function are $\mathbb{G}$-invariant under these transformations, and thus the critic is encouraged to learn a value function $V^{\pi}$ that is approximately $\mathbb{G}$-invariant, which in turn biases the actor toward a symmetric policy.
To mitigate the potential off-policy effect introduced by augmented samples under PPO, we follow the strategy of initializing the policy network with zero bias and a relatively large action variance, ensuring that the initial policy satisfies $\pi_{\theta}(g \,\triangleright\, \boldsymbol{s})\approx\pi_{\theta}(\boldsymbol{s})$ and thus provides sufficiently diverse gradient signals. This design allows both the original and augmented samples to contribute equally to policy updates, encouraging the policy to remain approximately equivariant throughout training~\cite{suleveraging}.

By explicitly leveraging symmetry through augmentation, the learning process is encouraged to explore state-action modalities more uniformly, which is expected to (i) improve sample efficiency by augmenting data without additional interaction, (ii) promote symmetry in locomotion trajectories across multiple command types, improving their controllability and practical applicability and (iii) enhance the symmetry of motion patterns themselves, leading to more natural and robust gait behaviors.

\begin{table}[tp]
	\centering
	\small
	\renewcommand{\arraystretch}{1.15}
  \setlength\tabcolsep{0.3mm}
  \caption{Symmetry transformations applied to the state space $\mathcal{S}$ and action space $\mathcal{A}$.}
	\label{tab:trans}
	\begin{tabular}{c|c|c} \whline{1pt}
        Trans. & Space & Vector \\ \whline{0.7pt}
        \multirow{4}{*}{$e$}
          & \multirow{3}{*}{$\mathcal{S}$}
          & \multirowcell{3}[0pt][c]{
              $[v_x^{{*}}, v_y^{{*}}, \omega^{{*}}, \Delta x, \Delta y, \Delta \psi, \sin\phi_\text{LF}, \cos\phi_\text{LF},$\\
              $\sin\phi_\text{LH}, \cos\phi_\text{LH},\sin{\phi_\text{RF}}, \cos\phi_\text{RF},\sin\phi_\text{RH}, $\\ 
              $\cos\phi_\text{RH},\sin\phi_\text{S}, \cos\phi_\text{S}, \text{cov}_\text{LF}, \text{cov}_\text{LH}, \text{cov}_\text{RF}, \text{cov}_\text{RH}]$
            } \\
             & &  \\ 
             & &  \\\cline{2-3} 
          & $\mathcal{A}$ & $[\omega_\text{LF}, \omega_\text{LH}, \omega_\text{RF}, \omega_\text{RH}, \omega_\text{S}]$ \\
        \whline{0.7pt}

        \multirow{4}{*}{$g_1$}
        & \multirow{3}{*}{$\mathcal{S}$}
          & \multirowcell{3}[0pt][c]{
              $[-v_x^{{*}}, v_y^{{*}}, -\omega^{{*}}, -\Delta x, \Delta y, -\Delta \psi, \sin\phi_\text{LH}, -\cos\phi_\text{LH},$\\
              $\sin\phi_\text{LF}, -\cos\phi_\text{LF},\sin{\phi_\text{RH}}, -\cos\phi_\text{RH},\sin\phi_\text{RF}, $\\ 
              $-\cos\phi_\text{RF},\sin\phi_\text{S}, \cos\phi_\text{S}, \text{cov}_\text{LH}, \text{cov}_\text{LF}, \text{cov}_\text{RH}, \text{cov}_\text{RF}]$
            } \\
             & &  \\ 
             & &  \\\cline{2-3} 
          & $\mathcal{A}$ & $[-\omega_\text{LH}, -\omega_\text{LF}, -\omega_\text{RH}, -\omega_\text{RF}, \omega_\text{S}]$ \\
        \whline{0.7pt}
        
        \multirow{4}{*}{$g_2$}
          & \multirow{3}{*}{$\mathcal{S}$}
          & \multirowcell{3}[0pt][c]{
              $[v_x^{{*}}, -v_y^{{*}}, -\omega^{{*}}, \Delta x, -\Delta y, -\Delta \psi, \sin\phi_\text{RF}, \cos\phi_\text{RF},$\\
              $\sin\phi_\text{RH}, \cos\phi_\text{RH},\sin{\phi_\text{LF}}, \cos\phi_\text{LF},\sin\phi_\text{LH}, $\\ 
              $\cos\phi_\text{LH},\sin\phi_\text{S}, -\cos\phi_\text{S}, \text{cov}_\text{RF}, \text{cov}_\text{RH}, \text{cov}_\text{LF}, \text{cov}_\text{LH}]$
            } \\
             & &  \\ 
             & &  \\\cline{2-3} 
          & $\mathcal{A}$ & $[\omega_\text{RF}, \omega_\text{RH}, \omega_\text{LF}, \omega_\text{LH}, -\omega_\text{S}]$ \\
        \whline{0.7pt}

        \multirow{4}{*}{$g_1\!\circ\! g_2$}
          & \multirow{3}{*}{$\mathcal{S}$}
          & \multirowcell{3}[0pt][c]{
              $[-v_x^{{*}}, -v_y^{{*}}, \omega^{{*}}, -\Delta x, -\Delta y, \Delta \psi, \sin\phi_\text{RH}, -\cos\phi_\text{RH},$\\
              $\sin\phi_\text{RF}, -\cos\phi_\text{RF},\sin{\phi_\text{LH}}, -\cos\phi_\text{LH},\sin\phi_\text{LF}, $\\ 
              $-\cos\phi_\text{LF},\sin\phi_\text{S}, -\cos\phi_\text{S}, \text{cov}_\text{RH}, \text{cov}_\text{RF}, \text{cov}_\text{LH}, \text{cov}_\text{LF}]$
            } \\
             & &  \\ 
             & &  \\\cline{2-3} 
          & $\mathcal{A}$ & $[-\omega_\text{RH}, -\omega_\text{RF}, -\omega_\text{LH}, -\omega_\text{LF}, -\omega_\text{S}]$ \\
        \whline{1pt}
	\end{tabular} 
\end{table}

\section{Experiments}\label{sec:exp}
In this section, we present experimental evaluations of the proposed framework. We first examine the locomotion trajectories generated by the robot under multiple gait commands, followed by the analysis of phase evolution under multi-gait command switching. We also conduct ablation studies on phase coverage reward and symmetry-based data augmentation to assess their contributions, with results evaluated in terms of both phase behavior and cumulative reward. 
Finally, we compare the proposed approach with the end-to-end control scheme to further demonstrate the effectiveness of the overall framework.
Visualizations of all hardware experiments are provided in the supplementary video.

\begin{table}[tp]
	\centering
	\footnotesize
	\renewcommand{\arraystretch}{1.15}
	\setlength\tabcolsep{2mm}
	\caption{Hyperparameter settings for implementation.}
	  \label{tab:para}
	  \begin{tabular}{cc} 
        \whline{1pt}	
         \textbf{Parameter} & \textbf{Value/Setting} \\
	\whline{0.5pt}
        Training episodes           & 100,000 \\
        Episode length              & 300 \\
        Batch size                  & 42,000 \\
        Number of epochs            & 20 \\
        Clip range                  & 0.2 \\
        Entropy coefficient         & 0.01 \\
        Discount factor ($\gamma$)  & 0.99 \\
        Learning rate               & 0.001 \\
        Units per layer             & [1024, 512, 256] \\
        Activation function         & ELU \\
        Leg joint parameters ($A$, $\theta_\mathrm{offset}$)   & $\pi/12, 31\pi/180$ \\
        Control interval ($\Delta t$) & 0.1 s\\
        Desired gait period ($T^*$) & 5 s\\
        Reward parameters ($\alpha, \beta, \zeta, \epsilon, \delta$) & $1, 1.5, 0.5, 2.2, 0.75$ \\
        Reward weights ($w_i$) & [0.1, 0.1, 0.1, 0.1, 0.1, 0.01]\\
		\whline{1pt}
	  \end{tabular} 
\end{table}

\begin{figure}[tp]
	\centering
	\includegraphics[width=1\linewidth]{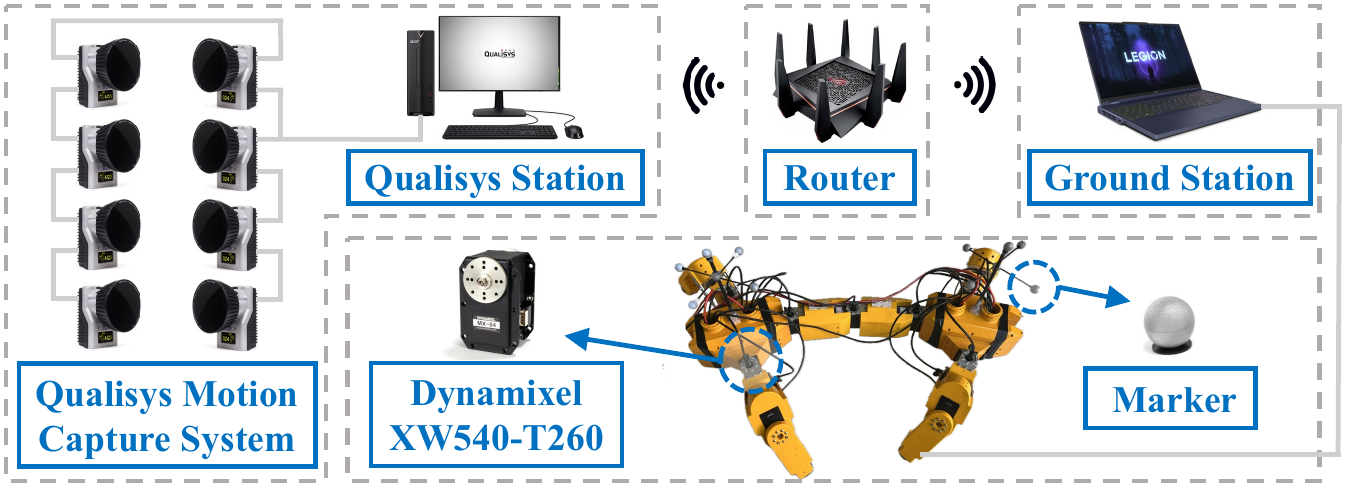}
    \caption{Schematic of the experimental platform.}
	\label{fig:court}
\end{figure}

\subsection{Implementation Details}
Both the policy and value networks are implemented as multi-layer perceptrons (MLPs) and optimized using PPO. The value network shares the same backbone as the policy network, differing only in its output dimension. The detailed hyperparameter settings are summarized in \tabref{tab:para}.  
At each episode reset, the phase values are uniformly initialized, while the gait categories are randomly shuffled and traversed sequentially, and a movement command is uniformly sampled for each category. MuJoCo~\cite{todorov2012mujoco} serves as the simulation environment due to its fast and accurate dynamics. For the sim-to-real transfer, the ground friction coefficient, mass distribution, and joint PID gains are carefully tuned to match the dynamics of the physical robot.

As illustrated in \figref{fig:court}, the experimental platform consists of a salamander-like robot, a Qualisys motion capture system, a Wi-Fi router, and a ground station. The robot's body parts are 3D-printed and actuated by Dynamixel XW540-T260 servomotors. The policy runs on the ground station, receiving real-time pose data from the Qualisys system via Wi-Fi and sending the computed motor commands back to the robot. The Qualisys system tracks reflective markers attached to the front and hind girdles to estimate their planar positions, which are used to establish the body coordinate frame.

\subsection{Learned Multi-Gait Behaviors}
\begin{figure}[tp]
	\centering
	\includegraphics[width=1\linewidth]{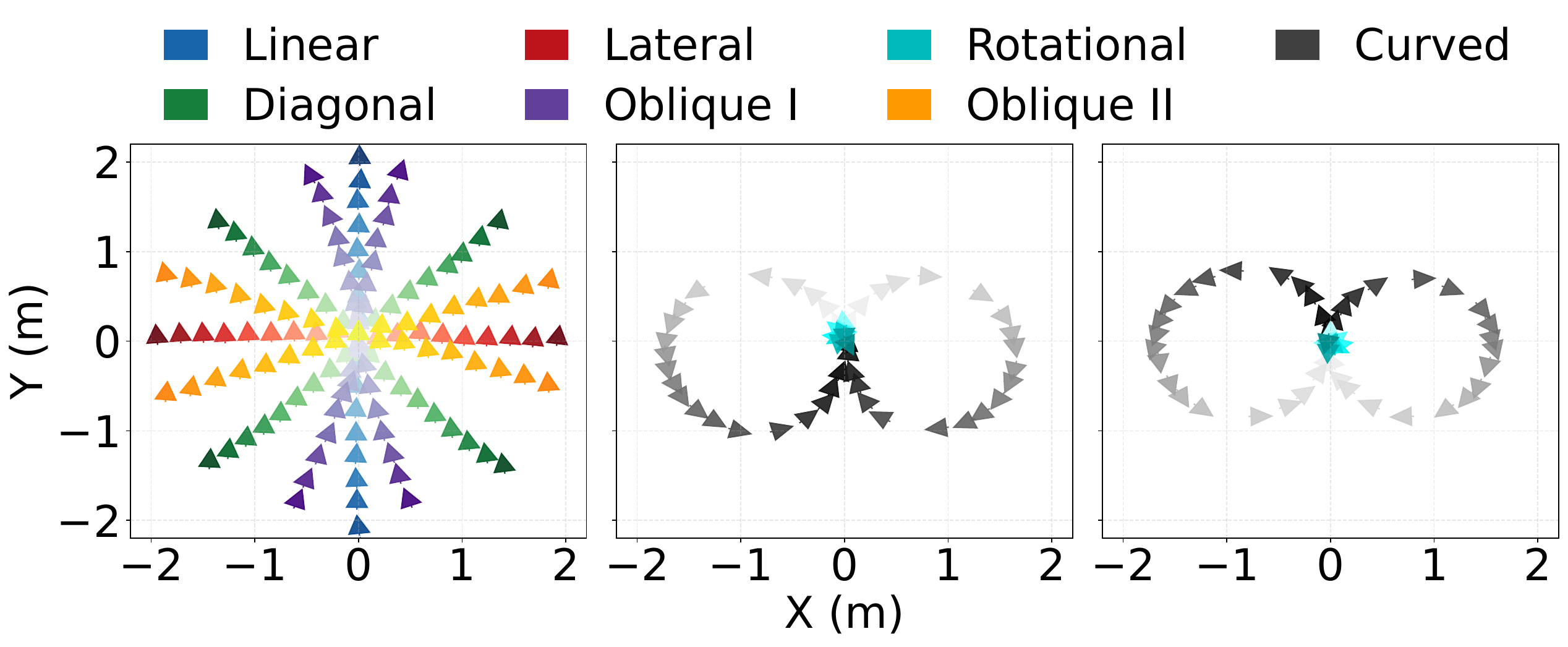}
	\caption{Simulated robot trajectories under seven gait categories (see \tabref{tab:gait_type}). Each trajectory is visualized as a sequence of arrows, with arrow orientation indicating body yaw and arrow color fading from light to dark to represent temporal progression.
    The rotational gaits are shown in the second and third subplots, corresponding to in-place counterclockwise and clockwise rotations.}
	\label{fig:traj}
\end{figure}

To evaluate the robot's capability to execute diverse gait commands after training, the learned policy is tested in simulation by executing each command and recording the resulting trajectory. Each command is executed for a sufficiently long duration, which may vary across commands, to obtain a clear trajectory. The resulting trajectories, corresponding to seven gait categories defined in \tabref{tab:gait_type}, are visualized in \figref{fig:traj}, with all of them starting from the origin. 

The results show that the robot successfully learns to follow commands involving different translational and rotational directions, thereby acquiring a diverse set of gait behaviors.
In the first subplot, the uniform arrow orientations reflect pure translational motion, whereas in the second and third subplots, the robot demonstrates the ability to turn with a finite radius and to rotate in place. Importantly, although previous studies have reported linear and curved locomotion gaits~\cite{crespi2013salamandra,horvat2017spine}, to the best of our knowledge, this is the first study to demonstrate that a salamander-like robot can achieve lateral, in-place rotational, and diagonal movements, thus expanding the known motion repertoire of such robots.
Moreover, as shown in \figref{fig:traj}, the trajectories of the same gait category exhibit strong central symmetry, and this symmetry is maintained across all gait types,
demonstrating that the data augmentation strategy successfully promotes task-level symmetry in learned behaviors.

\begin{figure}[tp]
	\centering
	\includegraphics[width=1\linewidth]{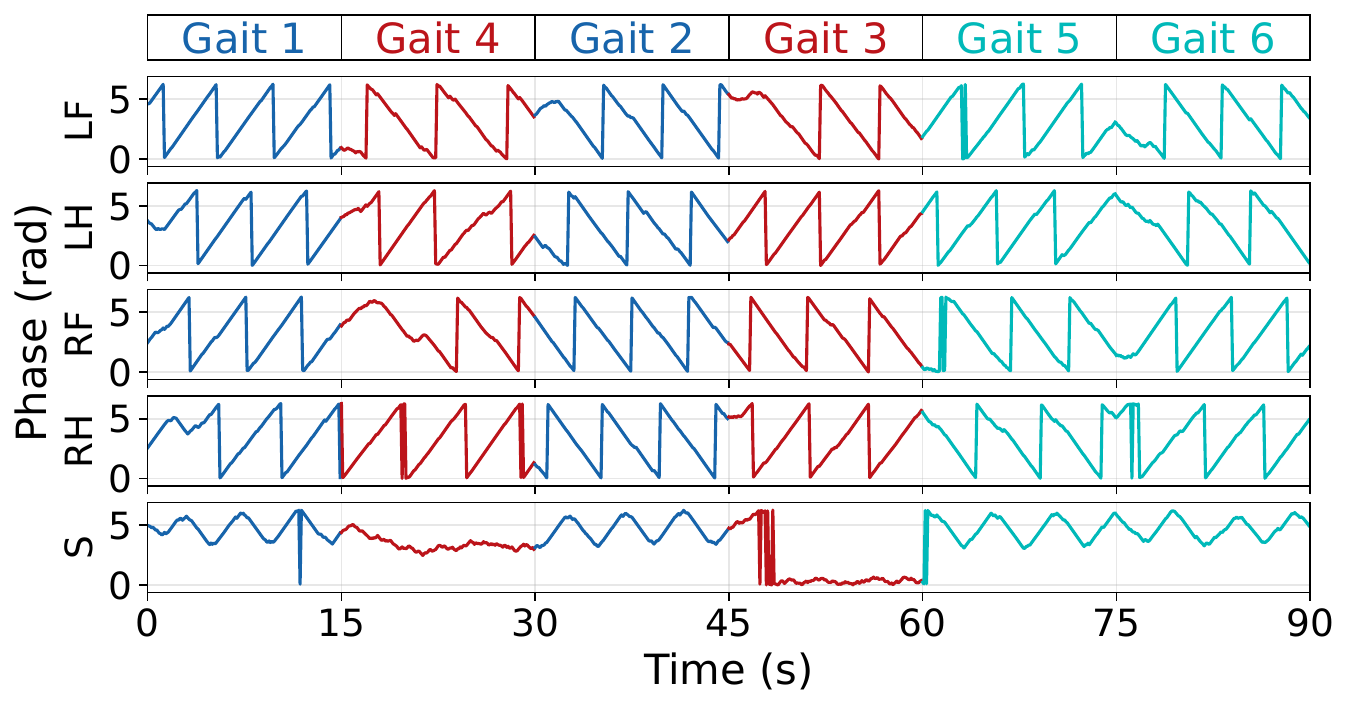}
	\caption{Phase evolution during gait transitions in real-world experiments (Gait 1 $\rightarrow$ 4 $\rightarrow$ 2 $\rightarrow$ 3 $\rightarrow$ 5 $\rightarrow$ 6 in \tabref{tab:gait_type}).
    Colors correspond to different gait categories in \tabref{tab:gait_type} and follow the same convention as in \figref{fig:traj}.}
	\label{fig:phase}
\end{figure}

\begin{table*}[t]
	\centering
	\small
	\renewcommand{\arraystretch}{1.15}
  \setlength\tabcolsep{1.05mm}
  \caption{Experimental comparison of cumulative rewards across 22 gait commands, including per-category mean and variance, between the proposed method and two ablations (w/o Aug.,w/o Cov.).}
	\label{tab:efficiency}
	\begin{tabular}{c|c|c|c|c|c|c|c|c|c|c|c|c|c|c|c|c|c|c} \whline{1pt}
        \multirow{3}*{Method} & \multicolumn{4}{c|}{Linear Locomotion} & \multicolumn{4}{c|}{Lateral Locomotion} & \multicolumn{4}{c|}{Rotational Locomotion} & \multicolumn{6}{c}{Curved Locomotion}\\ \cline{2-19}

        & \multicolumn{2}{c|}{Gait} & \multirow{2}*{$\mu_R$~$\uparrow$} & \multirow{2}*{$\sigma_R^2$~$\downarrow$} & \multicolumn{2}{c|}{Gait} & \multirow{2}*{$\mu_R$~$\uparrow$} & \multirow{2}*{$\sigma_R^2$~$\downarrow$} & \multicolumn{2}{c|}{Gait} & \multirow{2}*{$\mu_R$~$\uparrow$} & \multirow{2}*{$\sigma_R^2$~$\downarrow$} & \multicolumn{4}{c|}{Gait} & \multirow{2}*{$\mu_R$~$\uparrow$} & \multirow{2}*{$\sigma_R^2$~$\downarrow$} \\ \cline{2-3}\cline{6-7}\cline{10-11}\cline{14-17}

         & 1 & 2 & & & 3 & 4 & & & 5 & 6 & & & 7 & 8 & 9 & 10 & & \\ \whline{0.7pt}

         w/o Aug. & 9.18 & 13.31 & 11.24 & 8.51 & 7.65 & 6.74 & 7.20 & 0.41 & 7.10 & 12.19 & 9.64 & 12.95 & 5.94 & 9.93 & \textbf{14.32} & 10.14 & 10.08 & 11.71\\
         
         w/o Cov.  & 6.32 & 9.33 & 7.83 & 4.53 & 6.88 & 7.30 & 7.09 & \textbf{0.09} & 23.60 & 23.24 & 23.42 & \textbf{0.06} & 8.23 & 7.82 & 7.55 & 9.94 & 8.38 & \textbf{1.15}\\
         
         Ours     & \textbf{13.76} & \textbf{13.94} & \textbf{13.85} & \textbf{0.02} & \textbf{10.86} & \textbf{11.44} & \textbf{11.15} & 0.17 & \textbf{24.10} & \textbf{23.34} & \textbf{23.72} & 0.29 & \textbf{9.92} & \textbf{13.58} & 9.25 & \textbf{11.74} & \textbf{11.12} & 3.79\\\whline{1pt}

        \multirow{3}*{Method} & \multicolumn{6}{c|}{Diagonal Locomotion} & \multicolumn{6}{c|}{Oblique Locomotion \Rmnum{1}} & \multicolumn{6}{c}{Oblique Locomotion \Rmnum{2}}\\ \cline{2-19}

         & \multicolumn{4}{c|}{Gait} & \multirow{2}*{$\mu_R$~$\uparrow$} & \multirow{2}*{$\sigma_R^2$~$\downarrow$} & \multicolumn{4}{c|}{Gait} & \multirow{2}*{$\mu_R$~$\uparrow$} & \multirow{2}*{$\sigma_R^2$~$\downarrow$} & \multicolumn{4}{c|}{Gait} & \multirow{2}*{$\mu_R$~$\uparrow$} & \multirow{2}*{$\sigma_R^2$~$\downarrow$} \\ \cline{2-5}\cline{8-11}\cline{14-17}

         & 11 & 12 & 13 & 14 & & & 15 & 16 & 17 & 18 & & & 19 & 20 & 21 & 22 & & \\ \whline{0.7pt}

         w/o Aug. & 4.55 & 3.84 & 5.82 & 4.05 & 4.57 & 0.79 & 4.59 & 8.88 & 8.47 & 4.47 & 6.60 & 5.76 & 3.84 & 6.80 & 4.67 & 6.57 & 5.47 & 2.09\\
         
         w/o Cov.  & 3.74 & 2.84 & 2.79 & 3.32 & 3.17 & \textbf{0.20} & 6.87 & 6.50 & 6.25 & 6.67 & 6.57 & \textbf{0.07} & 4.51 & 5.07 & 4.72 & 4.95 & 4.81 & \textbf{0.06} \\
         
         Ours     & \textbf{5.92} & \textbf{7.01} & \textbf{6.98} & \textbf{6.31} & \textbf{6.55} & 0.28 & \textbf{10.35} & \textbf{11.31} & \textbf{10.06} & \textbf{11.20} & \textbf{10.73} & 0.38 & \textbf{9.62} & \textbf{9.71} & \textbf{10.18} & \textbf{9.79} & \textbf{9.82} & \textbf{0.06} \\
        
\whline{1pt}
	\end{tabular} 
\end{table*}

\subsection{Smooth Gait Transitions Under Changing Commands}
To further validate the policy's performance, we conduct experiments on the physical robot. The robot executes a sequence of commands, each lasting 15 seconds, in the following order: forward, right lateral, backward, and left lateral movements, followed by clockwise and counterclockwise rotations. The resulting phase evolution of the four legs and the spine is presented in \figref{fig:phase}.

Firstly, stable gait patterns with an approximate 5 s period emerge. During forward and backward walking, a typical trot gait~\cite{liu2024novel} can be observed: the four legs swing in aligned directions, with diagonal legs in-phase and bilateral legs in anti-phase, while the spine oscillates accordingly to increase stride length. During lateral motion, the spine bends toward one side, while the legs on the same side drive in the opposite direction. This counteracts forward velocity along the body x-axis and generates lateral velocity along the y-axis. For in-place rotations, the leg phases exhibit opposite progression patterns between the left and right sides. This phase arrangement generates opposing ground reaction forces that cancel translational motion while producing net yaw rotation of the body. Our results reveal that active spinal bending not only contributes to linear motion but also plays a crucial role in lateral walking and enhances the efficiency of in-place rotations. The learned policy fully exploits the structural features of the body and discovers novel motion patterns that have not been reported in the existing literature.

Besides, the robot exhibits symmetric gait patterns within each category. For instance, the phase trajectories of forward and backward walking follow the $g_{1}$ transformation, while those of left and right lateral walking correspond to $g_{2}$ transformation. This demonstrates that the symmetry-aware learning successfully induces symmetric movement patterns. 

Moreover, the phase trajectories show remarkably smooth and continuous transitions between gait patterns, despite the fact that the training regime involved only single-gait episodes with no explicit multi-command switching. Notably, the robot can adapt to any new gait command without abrupt jumps or phase discontinuities. This smoothness arises from two key design choices: (i) random initialization of leg and spine phases during training allows the agent to encounter diverse starting conditions, which promotes flexible adaptation to new commands, and (ii) the action space is parameterized in terms of frequency rather than absolute phase, with phase evolving as the integral of frequency over time. Consequently, phase evolution is inherently continuous, and abrupt phase resets are avoided. These mechanisms jointly enable robust, reliable, and smooth gait transitions, highlighting the generalization capability of the learned policy to handle sequential gait changes that were never explicitly demonstrated during training.

\subsection{Ablation on Phase Coverage Reward}
We train an ablation baseline (w/o Cov.) without using the phase coverage reward, while all other components and settings remain unchanged. To compare the two policies, the physical robot executes all 22 gait types for 25 seconds each, with three repetitions per gait. For evaluation, we compute the total undiscounted cumulative reward, along with the per-category mean and variance as performance metrics. The results are summarized in \tabref{tab:efficiency}.

The quantitative results in \tabref{tab:efficiency} show that our approach consistently outperforms the no-coverage baseline across all gait types. This improvement stems from the inclusion of the phase coverage reward, which encourages the robot to develop more dynamic gait patterns capable of covering longer distances and achieving larger turning angles. In contrast, without this reward, the learned gaits tend to be more static and suboptimal in accomplishing these objectives.

Beyond the final performance metrics, \figref{fig:rwd} provides further insight by comparing the training reward curves. Specifically, \figref{fig:rwd}(a) reports the total reward of the full method (Ours), the total reward with the coverage term removed at evaluation time (Ours (-Cov.)), and the policy trained entirely without the coverage reward (w/o Cov.). Notably, 
the full method (Ours) consistently achieves higher returns than the baseline trained without the phase coverage reward (w/o Cov.).
More importantly, even after removing the coverage term from the total reward at evaluation time, Ours (-Cov.) still outperforms the w/o Cov. baseline by a clear margin of about 2 reward units after convergence.
This observation indicates that the performance gain cannot be attributed solely to the numerical contribution of the coverage reward itself. Instead, the phase coverage term encourages broader exploration of the leg phase space during learning, leading the policy to discover more effective gait patterns that translate into improved overall locomotion performance.

\newcommand{\addFig}[1]{\includegraphics[width=0.49\linewidth]{figs/#1.pdf}}
\begin{figure}[t]
  \centering
  \small
  \setlength\tabcolsep{0.4mm}
  \renewcommand\arraystretch{1.2}
  \begin{tabular}{cc}
  \addFig{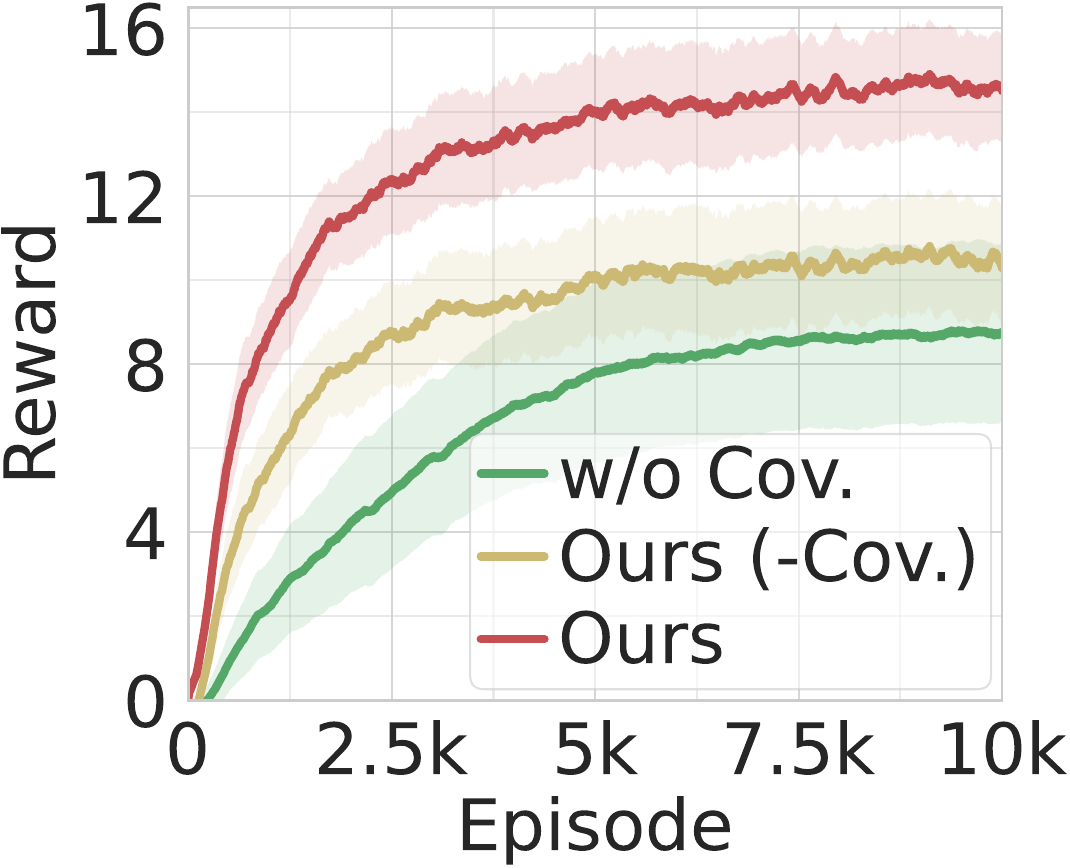} & \addFig{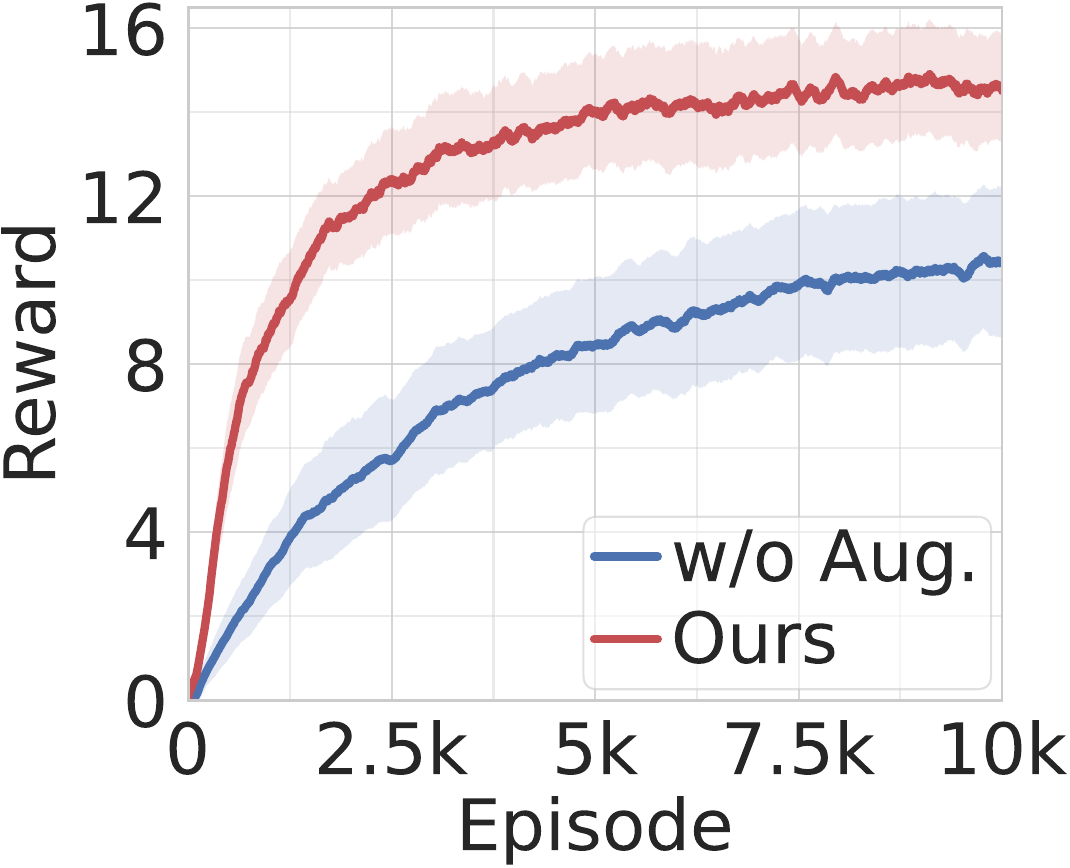} \\
  (a) & (b) \\
  \end{tabular}
  \caption{
    Training reward curves for ablation studies on phase coverage reward (a) and symmetry-based data augmentation (b).
    The solid lines represent the average value of every 500 episodes, and the shaded regions depict the mean $\pm$ half a standard deviation.
}
  \label{fig:rwd}
\end{figure}

\begin{figure}[tp]
	\centering
	\includegraphics[width=1\linewidth]{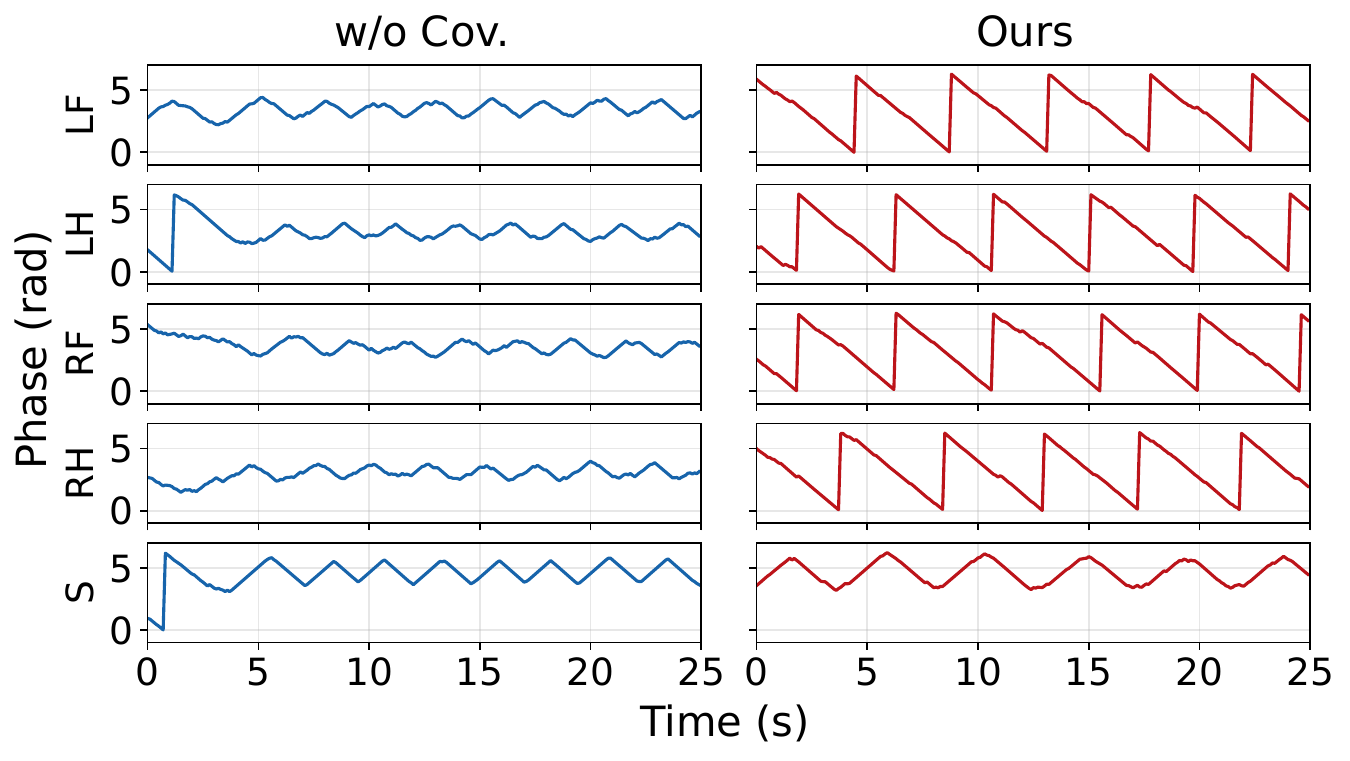}
    \caption{Experimental phase trajectories under backward command, comparing our method against the baseline without phase coverage reward (w/o Cov.).}
	\label{fig:cov}
\end{figure}

 \figref{fig:cov} compares the phase trajectories of the five body parts under the backward command (Gait 2 in \tabref{tab:gait_type}).  
The results reveal a striking difference: for the full framework, the leg phases exhibit clear periodicity and span nearly full $2\pi$ range after stabilization, while the spine phase, though oscillating twice within $[\pi, 2\pi]$, produces motion that is effectively equivalent to spanning the entire $2\pi$ range after the cosine mapping. Although the phase coverage reward does not include the spine phase, the spine achieves the desired motion by learning to coordinate with the limbs during training. In contrast, the no-coverage variant produces leg oscillations of less than $0.5\pi$, leading to relatively static limb behavior. This phenomenon highlights that incorporating phase coverage reward encourages dynamic and well-distributed gait patterns.

\subsection{Ablation on Symmetry-Based Data Augmentation}
To assess the effect of incorporating morphological symmetry, we train an ablation baseline (w/o Aug.) without using augmented samples, while keeping all other components and settings identical. 
\figref{fig:rwd}(b) compares the corresponding training reward curves. The results show that our full method consistently converges to a substantially higher return than the w/o Aug. baseline, with a gap of approximately 4 reward units after convergence. This clear separation in final performance demonstrates the effectiveness of symmetry-based data augmentation in improving learning outcomes, rather than merely accelerating early-stage training.

The testing results are also presented in \tabref{tab:efficiency}. As indicated in \tabref{tab:efficiency}, the use of symmetry-based data augmentation leads to a significant improvement in the cumulative rewards for most commands. The rewards show not only higher overall values but also considerably lower variances within the same category of gait commands. For instance, in the diagonal locomotion gait category, symmetry-based data augmentation increases the rewards for individual commands, resulting in a rise of the mean reward from 4.57 to 6.55. At the same time, the variance across commands decreases considerably, from 0.79 without augmentation to 0.28 with it. This result suggests that symmetry-based data augmentation fosters not only improved performance but also more consistent and symmetric movements across different gait commands.

\figref{fig:symm} illustrates the robot trajectories observed in simulation and the corresponding phase evolutions for four symmetric gait commands (Gait 11-14, as defined in \tabref{tab:gait_type}), all of which belong to the diagonal locomotion category. Without symmetry-based data augmentation, the robot's movements exhibit a certain degree of symmetry in the forward-left (Gait 11) and backward-left (Gait 12) directions, yet the trajectories are not symmetric across the left and right sides. The backward-right (Gait 13) and forward-right (Gait 14) directions deviate significantly from the intended diagonal directions, and the robot's yaw orientation also rotates to some extent. In addition, the gaits are irregular and fail to form a consistent pattern. In contrast, using symmetry-based data augmentation, our approach ensures that the robot moves along a nearly diagonal path without heading rotation ($\omega\approx0$), and the resulting gaits across all four directions are highly symmetrical. These phenomena demonstrate the effectiveness of the proposed symmetry-based augmentation.

\renewcommand{\addFig}[1]{\includegraphics[width=0.236\linewidth]{figs/symm/#1.jpg}}
\newcommand{\addpdf}[1]{\includegraphics[width=0.236\linewidth]{figs/symm/#1.pdf}}
\begin{figure}[t]
  \centering
  \small
  \setlength\tabcolsep{0.4mm}
  \renewcommand\arraystretch{0.9}
  \begin{tabular}{cccc}
  \hspace{0.15em} Gait 11 & Gait 14\hspace{0.15em} &Gait 11 & Gait 14 \\
  \hspace{0.15em}\addpdf{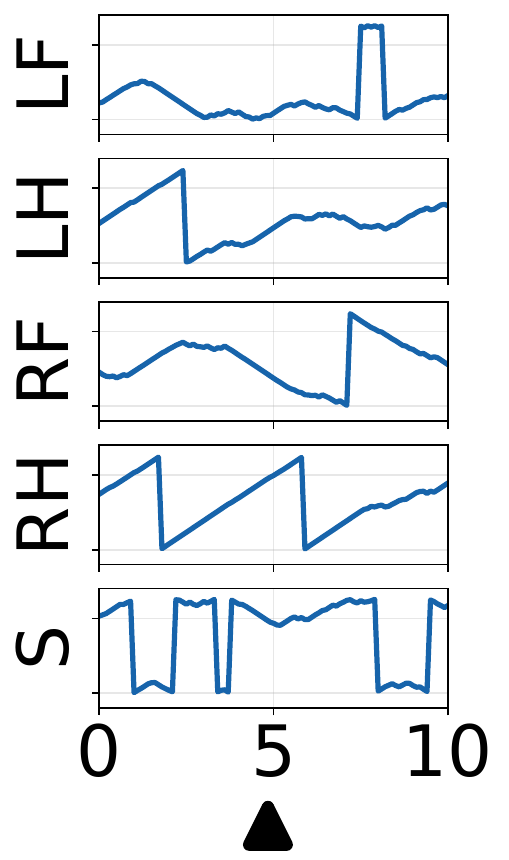} & \addpdf{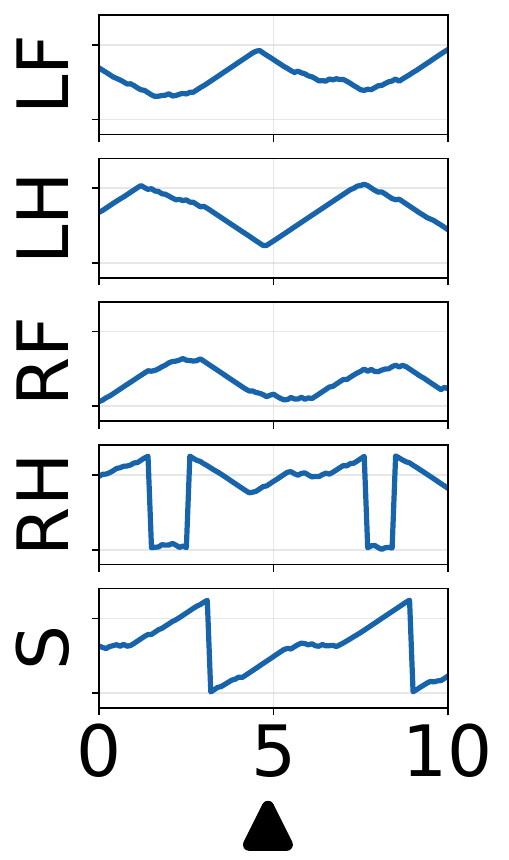}\hspace{0.15em} & \hspace{0.15em}\addpdf{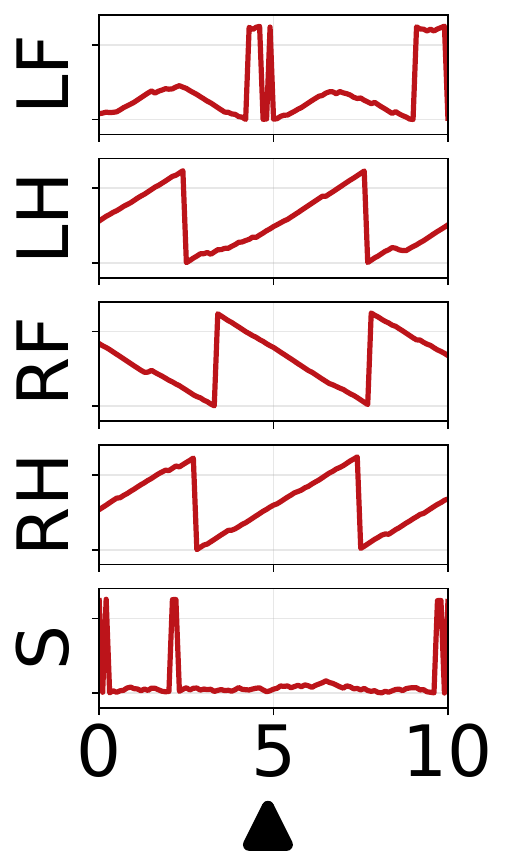} & \addpdf{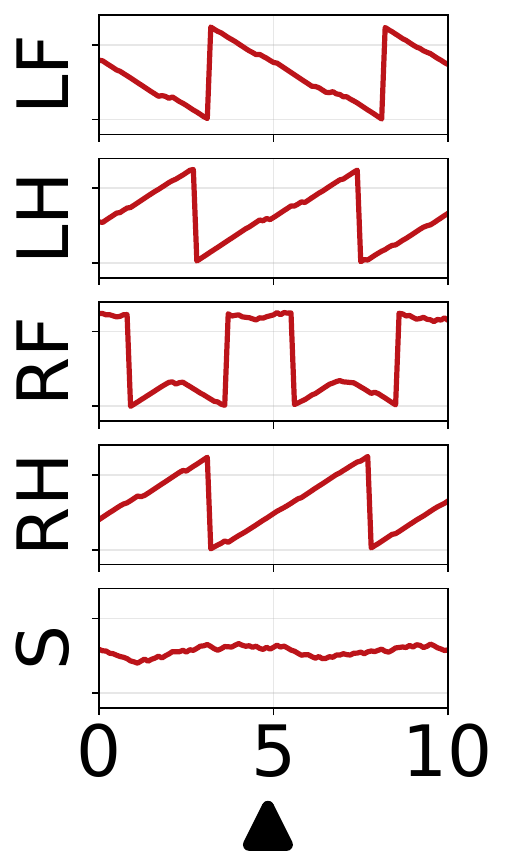}\hspace{0.15em} \\
  \hspace{0.15em}\addFig{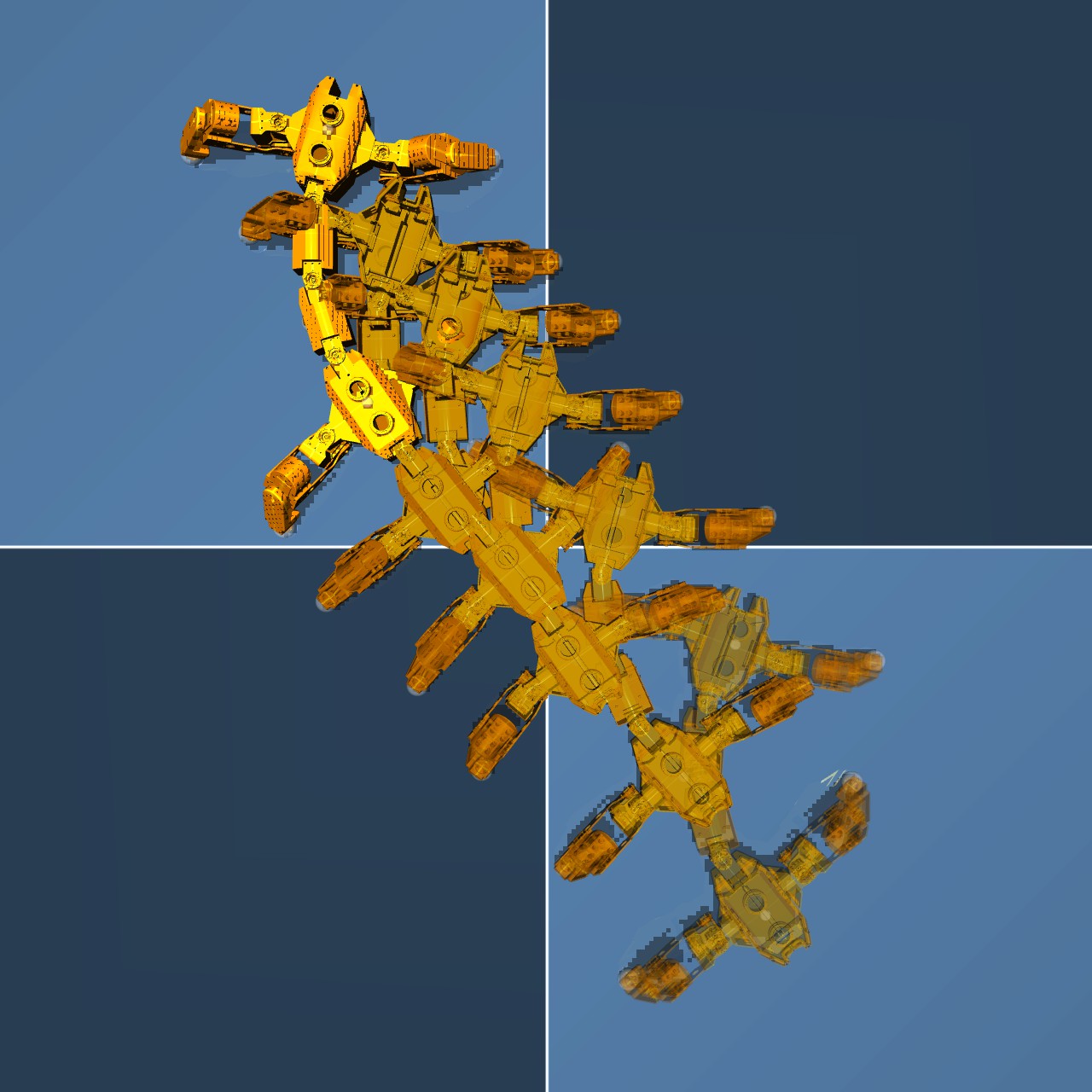} & \addFig{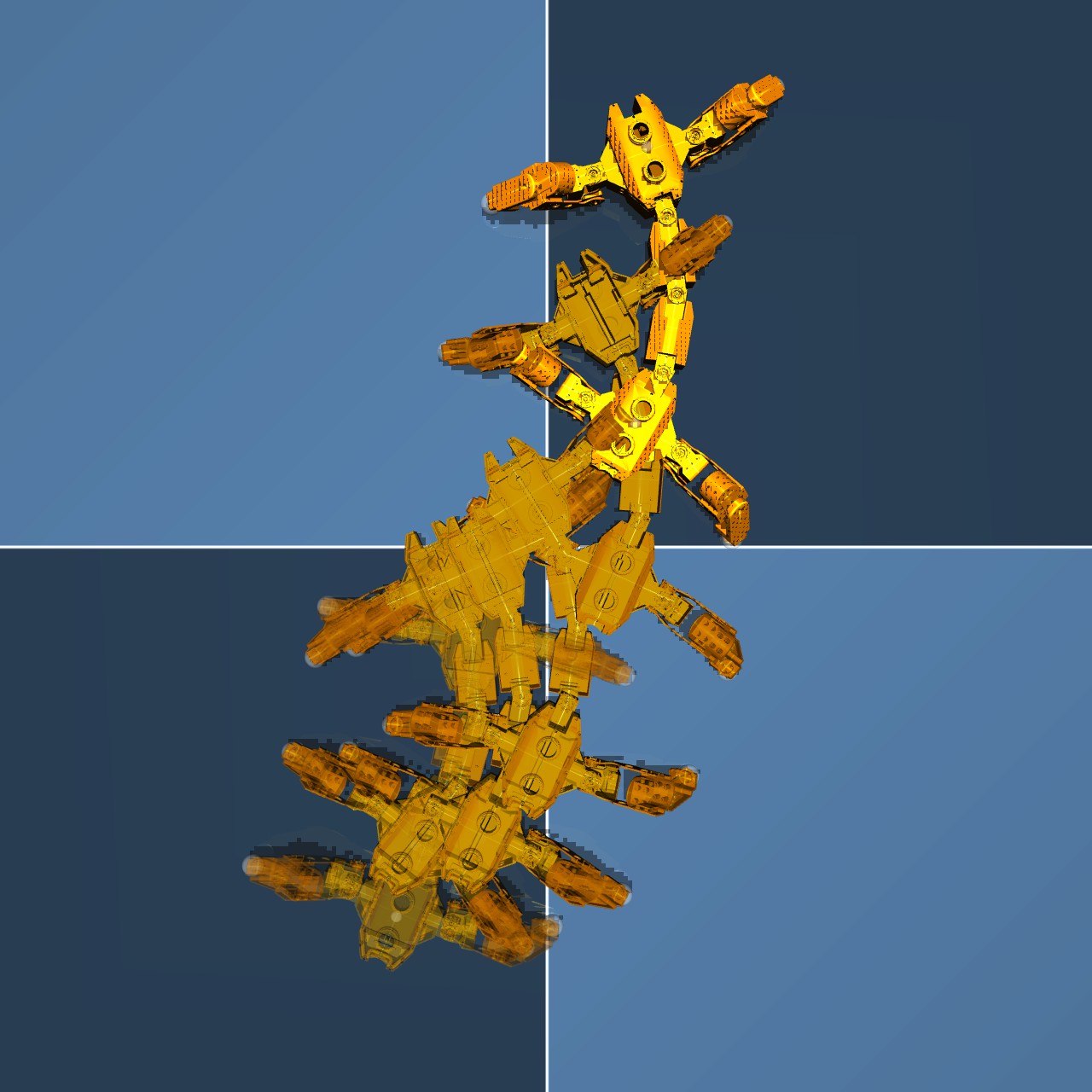}\hspace{0.15em} & \hspace{0.15em}\addFig{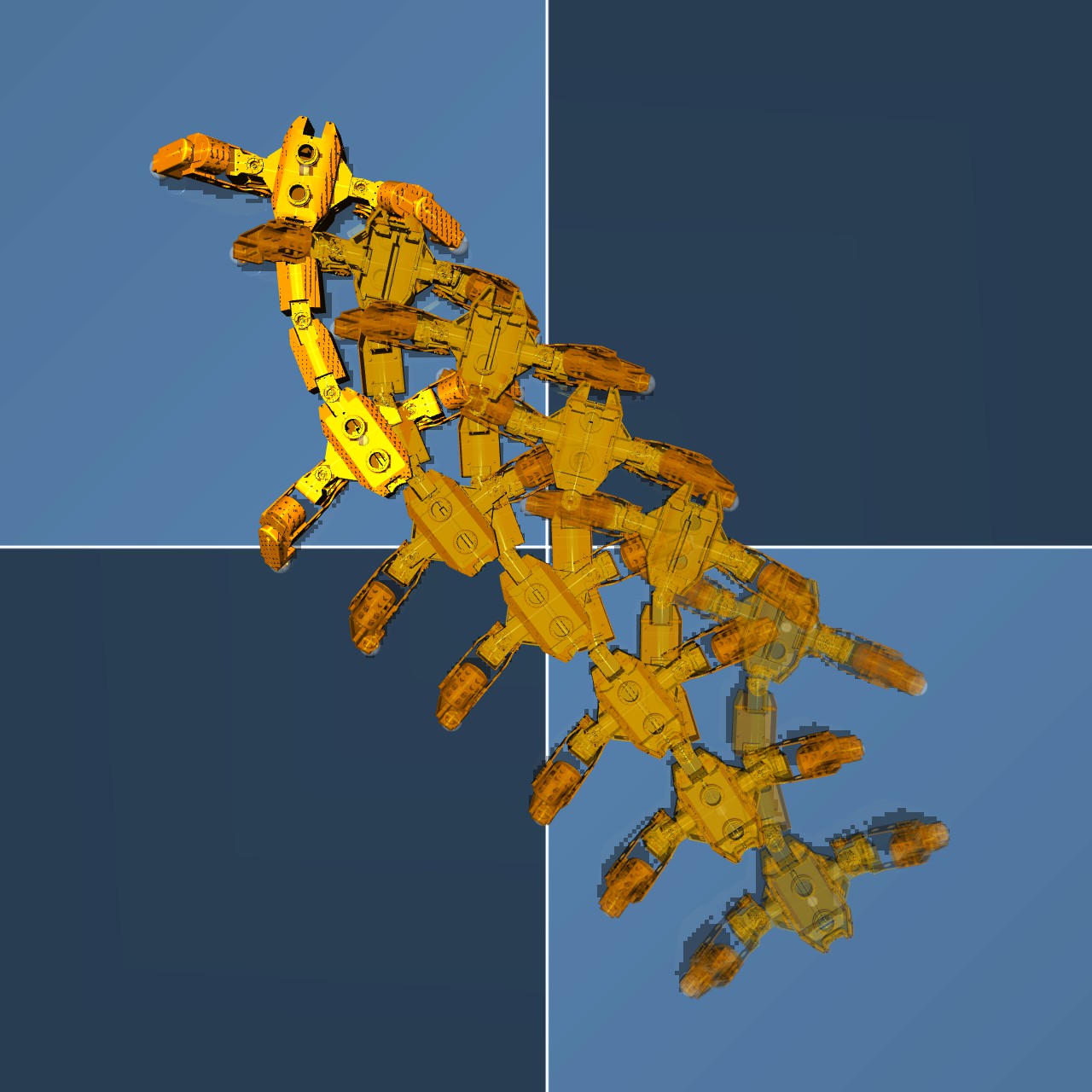} & \addFig{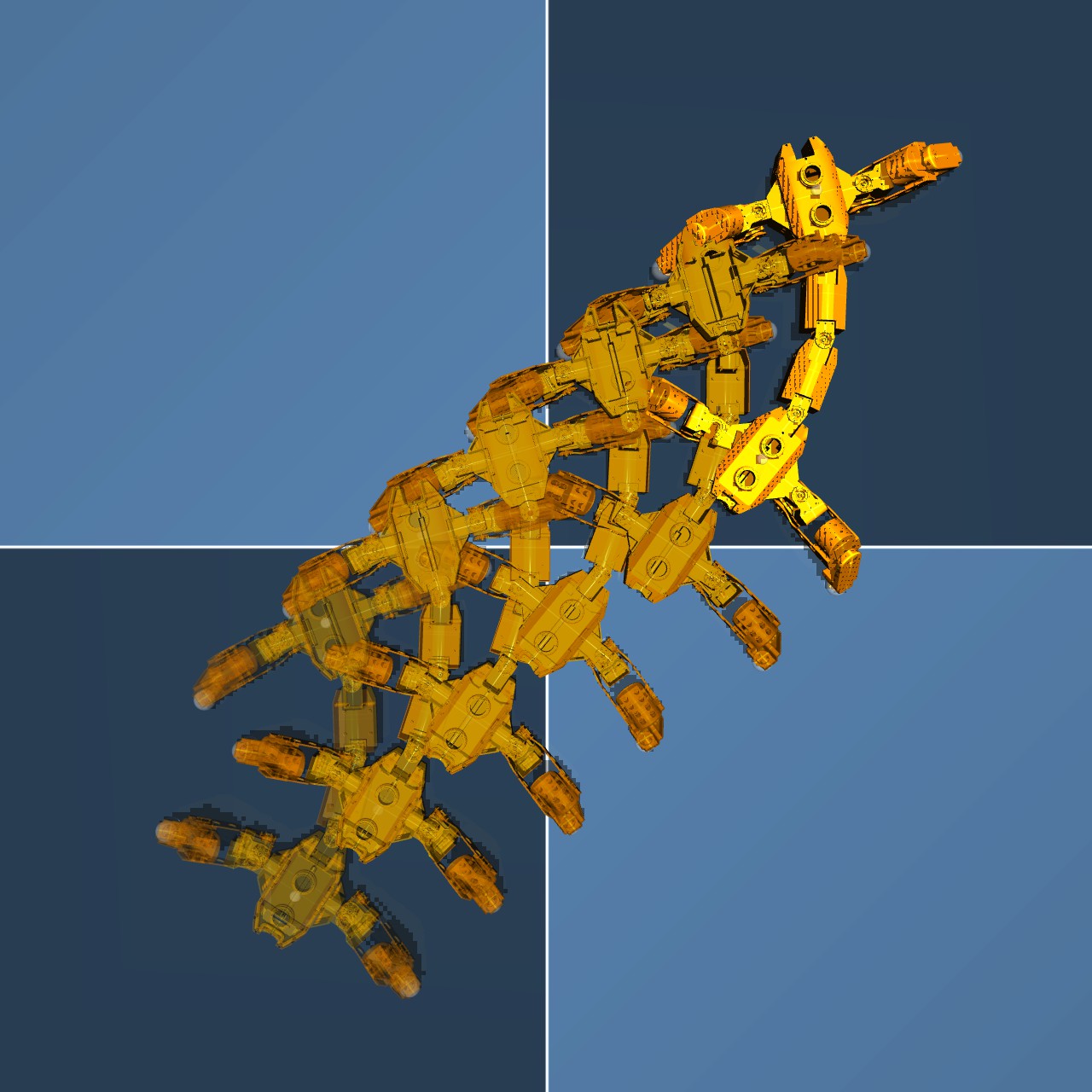}\hspace{0.15em} \\
  
  \hspace{0.15em}\addFig{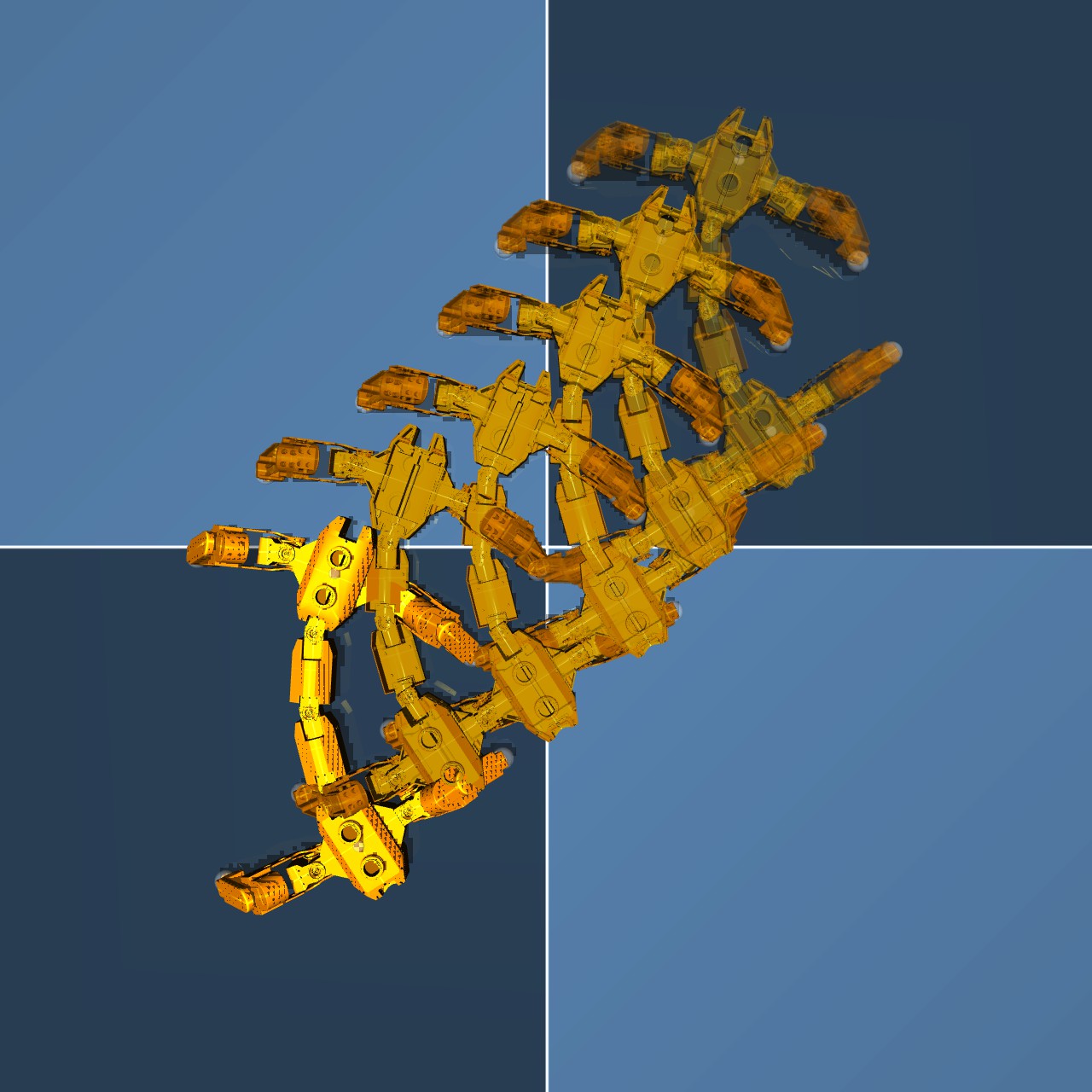} & \addFig{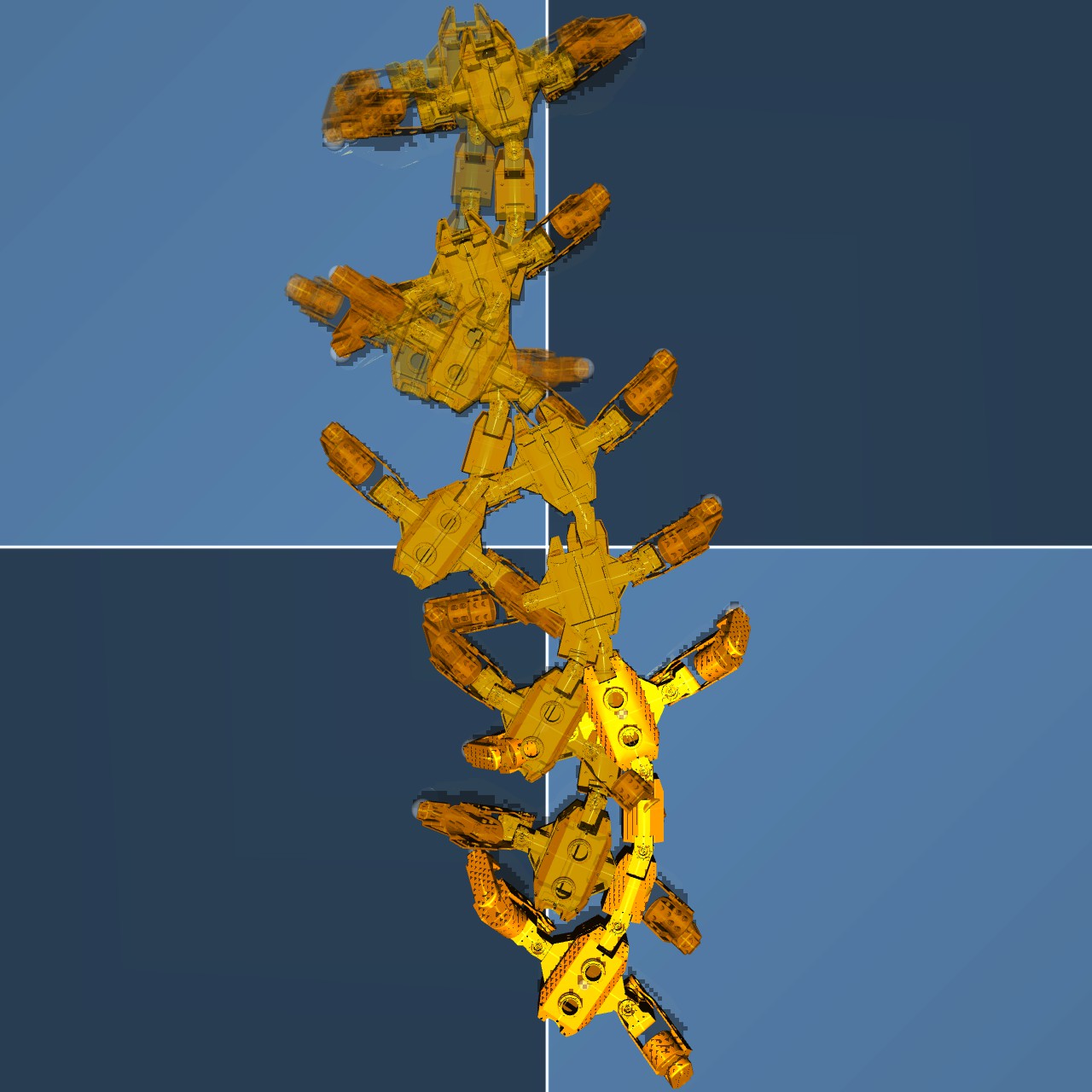}\hspace{0.15em} & \hspace{0.15em}\addFig{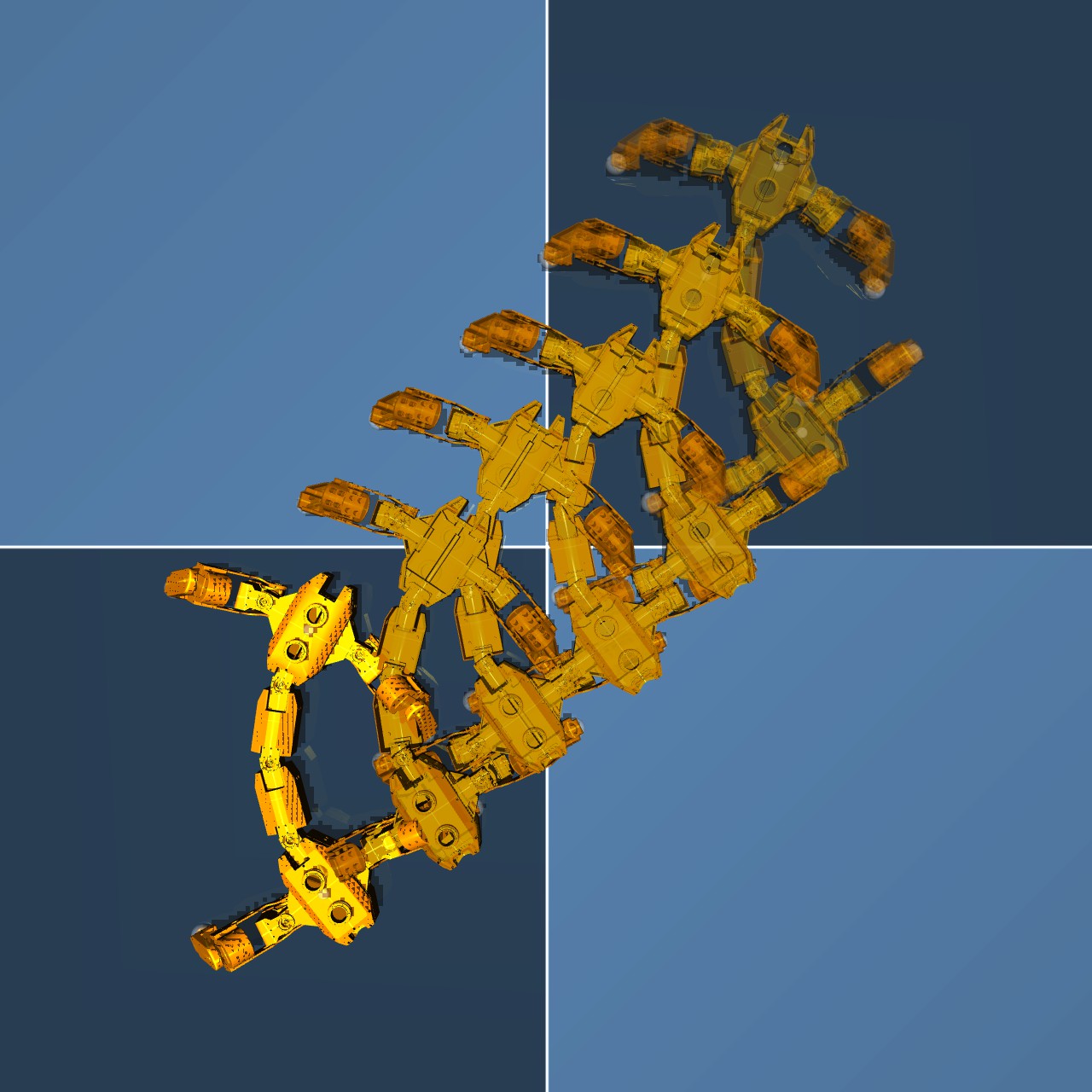} & \addFig{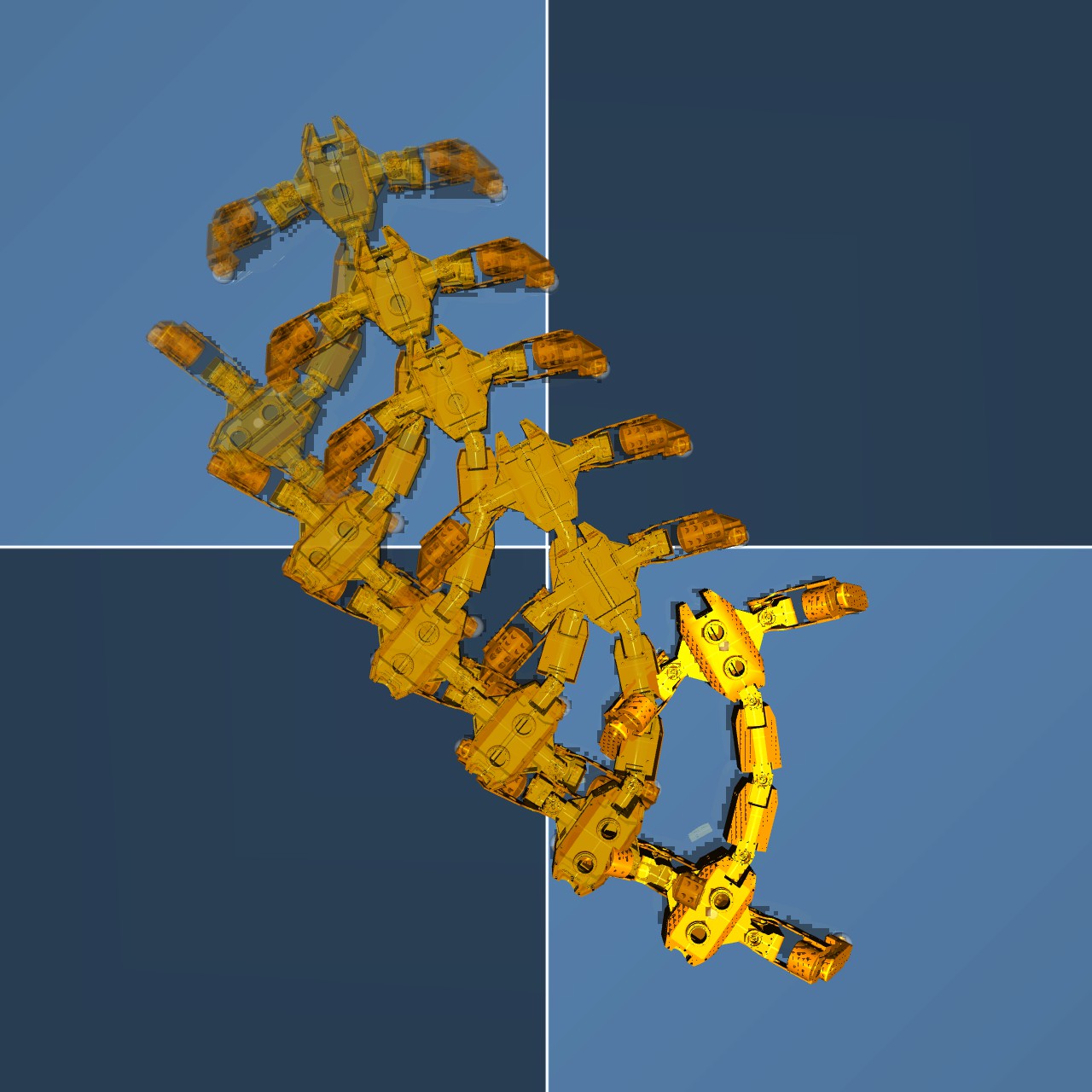}\hspace{0.15em} \\
  \hspace{0.15em}\addpdf{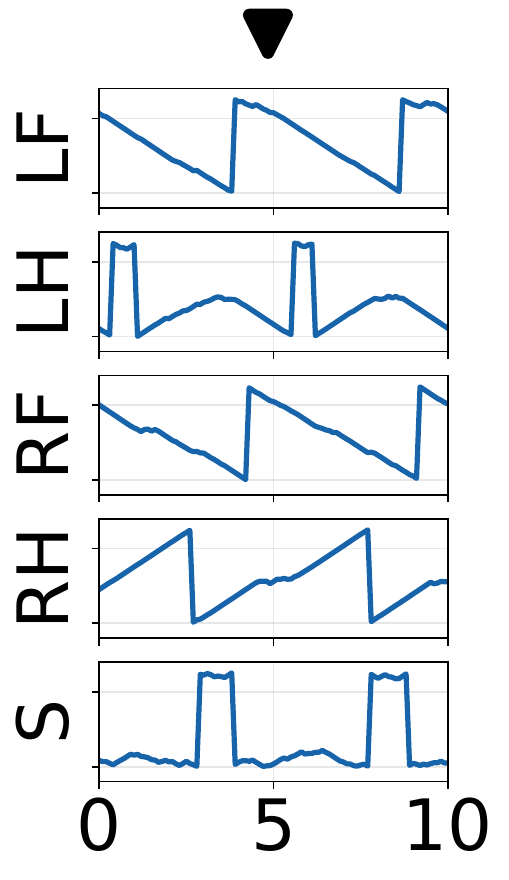} & \addpdf{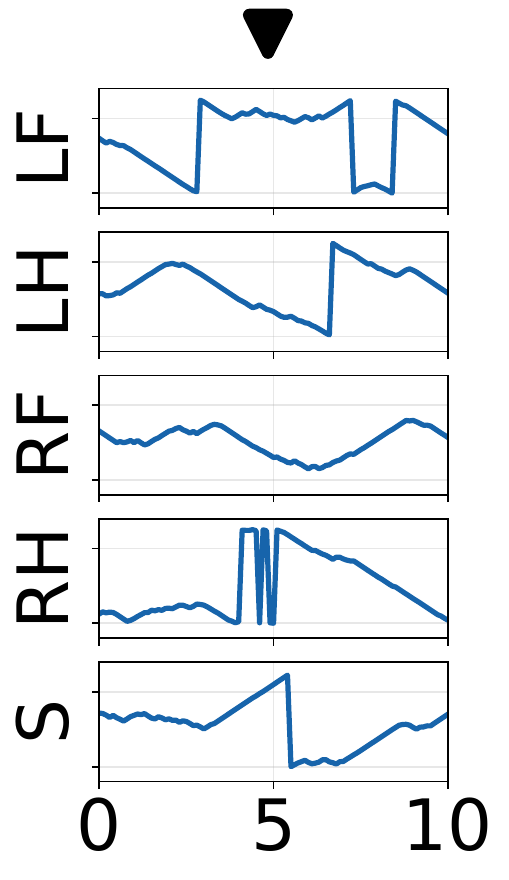}\hspace{0.15em} & \hspace{0.15em}\addpdf{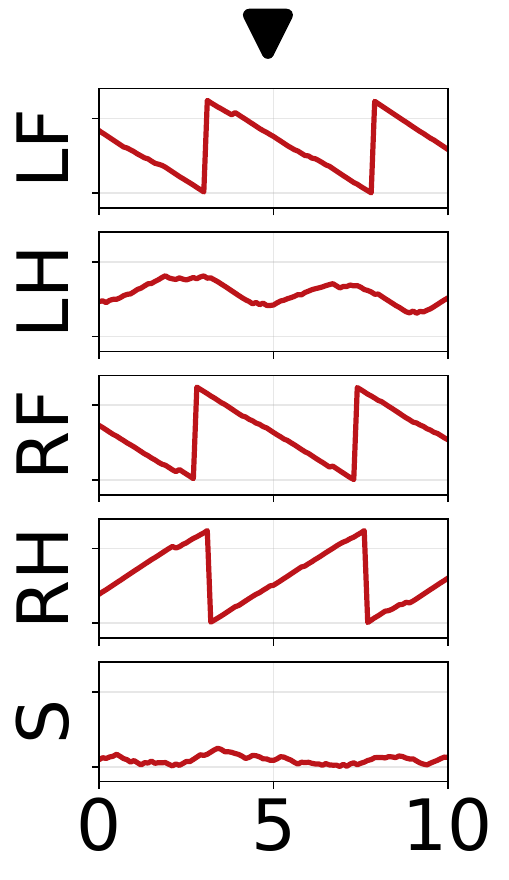} & \addpdf{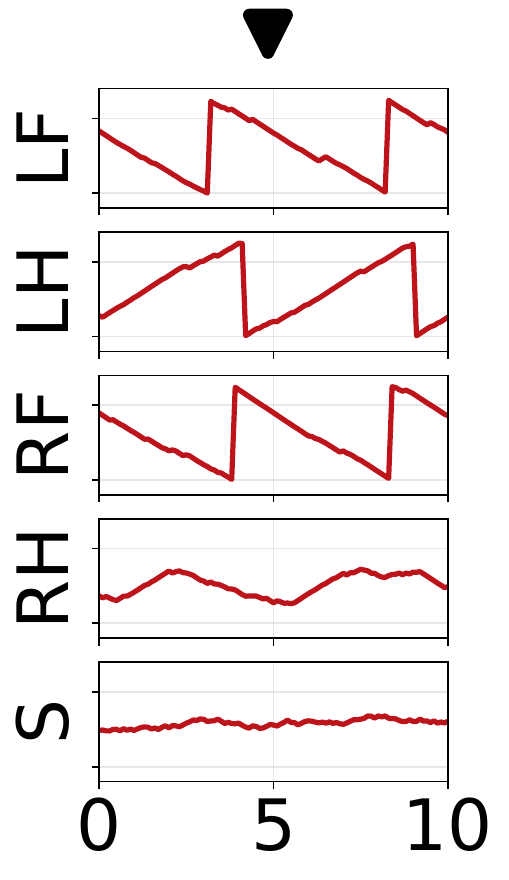}\hspace{0.15em} \\
  \hspace{0.15em} Gait 12 & Gait 13\hspace{0.15em} &Gait 12 & Gait 13 \\
  \multicolumn{2}{c}{(a) w/o Aug.} & \multicolumn{2}{c}{(b) Ours}\\
  \end{tabular}
  \caption{
    Robot trajectories in simulation and the corresponding phase evolution plots for four symmetric gait commands (Gaits 11--14 in \tabref{tab:gait_type}), all belonging to the diagonal locomotion category. The phase plots illustrate 10 s of evolution, with time on the horizontal axis and phase on the vertical axis.
}
  \label{fig:symm}
\end{figure}

\subsection{Comparison with the End-to-End Control Scheme}
To evaluate the effectiveness of the proposed phase-based multi-gait learning framework, a comparison is conducted against an end-to-end (E2E) reinforcement learning baseline. Unlike the proposed method, which operates in a phase velocity space coupled with a trajectory generator, the E2E baseline directly outputs a 15-dimensional joint-angle command at each control step. The action bounds are set to match those of the trajectory generator to ensure a fair comparison. To mitigate excessive high-frequency actuation, an additional action smoothing term is introduced into the optimization objective~\cite{chen2025learning}. With this modification, the E2E policy is able to converge to its best achievable performance during training.

\begin{table}[t]
	\centering
	\small
	\renewcommand{\arraystretch}{1.15}
  \setlength\tabcolsep{1.05mm}
  \caption{Quantitative comparison with an end-to-end (E2E) control scheme in simulation for translational and in-place rotational motions (mean $\pm$ standard deviation).
  }
	\label{tab:e2e}
	\begin{tabular}{c|c|c|c|c|c} \whline{1pt}
    \multicolumn{6}{l}{Linear, Lateral, and Diagonal Locomotion} \\\whline{1pt}
        \multirow{2}*{Gait} & \multicolumn{2}{c|}{Deviation Angle ($^\circ$)~$\downarrow$} & \multirow{2}*{Gait} & \multicolumn{2}{c}{Deviation Angle ($^\circ$)~$\downarrow$}\\ \cline{2-3} \cline{5-6}
        & E2E & Ours & & E2E & Ours \\ \whline{0.7pt}
        1 & 36.94$\pm$31.80 & \textbf{8.49$\pm$5.36} & 2 & 22.12$\pm$16.15 & \textbf{7.14$\pm$6.57} \\
        3 & 10.13$\pm$16.76 & \textbf{4.68$\pm$2.08} & 4 & 26.24$\pm$2.18 & \textbf{3.12$\pm$2.16} \\
        11 & 58.20$\pm$25.03 & \textbf{3.13$\pm$3.24} & 12 & 72.68$\pm$27.36 & \textbf{3.01$\pm$4.24}\\
        13 & 13.26$\pm$18.44 & \textbf{2.90$\pm$1.67} & 14 & 9.82$\pm$2.51 & \textbf{2.62$\pm$1.35}\\ \whline{1pt}
        \multicolumn{6}{l}{Rotational Locomotion} \\\whline{1pt}
        \multirow{2}*{Gait} & \multicolumn{2}{c|}{Deviation Distance (m)~$\downarrow$} & \multirow{2}*{Gait} & \multicolumn{2}{c}{Deviation Distance (m)~$\downarrow$}\\ \cline{2-3} \cline{5-6}
        & E2E & Ours & & E2E & Ours \\ \whline{0.7pt}
        5 & 0.72$\pm$0.35 & \textbf{0.16$\pm$0.11} & 6 & 0.65$\pm$0.33 & \textbf{0.17$\pm$0.08} \\\whline{1pt}
	\end{tabular}
\end{table}

The evaluation is conducted in simulation over 10 long-horizon rollouts, with each rollout lasting 100 s.
For translational motions-including linear, lateral, and diagonal locomotion-the deviation angle between the commanded direction and the actual displacement is measured. For in-place rotational commands, the deviation distance from the rotation center is used as the evaluation metric. Quantitative results are summarized in \tabref{tab:e2e}.

As shown in the table, the proposed method consistently achieves substantially smaller deviations across all tested gaits. 
For translational locomotion, the E2E baseline exhibits large directional errors, with deviation angles often exceeding 30$^\circ$-70$^\circ$ for several gaits, whereas the proposed method maintains deviations within only a few degrees (typically below 5$^\circ$). A similar trend is observed for in-place rotational motions: the E2E controller drifts noticeably from the rotation center, with mean deviation distances above 0.6 m, while the proposed method reduces this error to approximately 0.16-0.17 m.

These results indicate that, although the E2E controller can generate motion, its long-term behavior is significantly less aligned with the commanded motion objectives. This discrepancy becomes increasingly pronounced over extended execution horizons.
A key reason for this difference is that the E2E baseline does not incorporate gait-related auxiliary rewards, nor does it exploit phase or a dedicated trajectory generator.
As a result, the learned policy fails to develop coherent and repeatable gait patterns. The resulting motions are irregular and highly variable, which not only degrades tracking accuracy but also raises safety concerns. Consequently, despite successful convergence in simulation, the E2E controller is deemed unsuitable for deployment on real hardware.
In contrast, the proposed method, by combining phase with a trajectory generator, produces consistent, physically realizable gaits with higher tracking accuracy, suitable for real-world deployment.

\section{Conclusions and Future Work}\label{sec:conclusion}
This paper proposes a phase-based multi-gait learning framework for a salamander-like robot that enables the acquisition of a diverse repertoire of gaits without using reference motions. Each body part is controlled by a phase variable, while a phase coverage reward encourages exploration of the leg phase space. Morphological symmetry is incorporated through data augmentation, enhancing sample efficiency and promoting symmetry in both motion and task execution. Extensive experiments verified that the robot can master a wide range of dynamic and symmetric gaits, demonstrating the effectiveness of the framework.

Future work will explore augmenting the action space with joint-angle amplitudes and extending the proposed approach to other bio-inspired robotic platforms to further validate its generality across different morphologies.
Additionally, incorporating perceptual modalities such as vision into the framework will enable higher-level tasks such as obstacle avoidance and autonomous navigation, further advancing the autonomy of legged robotic systems.

\bibliographystyle{IEEEtran}
\bibliography{ref}

\end{document}